%% file: main.tex
\documentclass[10pt]{article} % For LaTeX2e
\usepackage[accepted]{tmlr}
\input{math_commands.tex}

\usepackage{hyperref}
\usepackage{url}
\usepackage{amsmath}
\usepackage{wrapfig}

\title{Text Rationalization for Robust Causal Effect Estimation}

\author{\name Lijinghua Zhang \email lijinghz@uci.edu \\
      \addr Department of Statistics\\
      University of California, Irvine
      \AND
      \name Hengrui Cai \email hengrc1@uci.edu \\
      \addr Department of Statistics\\
      University of California, Irvine
      }

\def\month{10}  % Insert correct month for camera-ready version
\def\year{2026} % Insert correct year for camera-ready version
\def\openreview{\url{https://openreview.net/forum?id=m8Vtjiu7SL}} % Insert correct link to OpenReview for camera-ready version

\input{preamble}

\begin{document}

\maketitle

\begin{abstract}
Recent advances in natural language processing have enabled the increasing use of text data in causal inference, particularly for confounding adjustment in treatment effect estimation. 
Although high-dimensional text can encode rich contextual information, it also poses unique challenges for causal identification and estimation. 
In particular, the positivity assumption, which requires sufficient treatment overlap across confounder values, is often violated at the observational level, when massive text is represented in feature spaces. 
Redundant or spurious textual features inflate dimensionality, producing extreme propensity scores, unstable weights, and inflated variance in effect estimates. 
We address these challenges with Confounding-Aware Token Rationalization (CATR), a framework that \rev{learns a sparse adjustment representation of the text using a residual-independence diagnostic to encourage the preservation of confounding information for unconfoundedness}. 
By discarding irrelevant texts while retaining key signals, CATR mitigates observational-level positivity violations and stabilizes downstream causal effect estimators. 
\revv{Semi-synthetic experiments demonstrate improved causal effect accuracy and stability, and a real-world MIMIC-III study shows improved empirical overlap and estimator stability.
Together, they illustrate the effectiveness and interpretability of our proposed method.}
\end{abstract}

\input{Sections/1-introduction}

\input{Sections/2-notation}

\input{Sections/3-hsic}
\input{Sections/3-method}

\input{Sections/4-theoretical}

\input{Sections/5-experiment}

\input{Sections/6-conclusion}

\bibliography{main}
\bibliographystyle{tmlr}

\appendix

\numberwithin{figure}{section}
\numberwithin{table}{section}

\input{Sections/8-appendix}

\end{document}

%% file: math_commands.tex
\usepackage{amsmath,amsfonts,bm}

\def\eqref#1{equation~\ref{#1}}
\def\1{\bm{1}}

\DeclareMathAlphabet{\mathsfit}{\encodingdefault}{\sfdefault}{m}{sl}
\SetMathAlphabet{\mathsfit}{bold}{\encodingdefault}{\sfdefault}{bx}{n}

\newcommand{\E}{\mathbb{E}}

\newcommand{\Var}{\mathrm{Var}}

%% file: preamble.tex
\usepackage{graphicx}
\usepackage{enumerate}
\usepackage{enumitem}
\usepackage{amsthm}

\usepackage{algorithm}
\usepackage{algorithmic}

\usepackage{booktabs}
\usepackage{multirow}

\usepackage{xcolor}

\newif\ifrevisions
\revisionsfalse
\definecolor{revcolor}{RGB}{0,0,200}
\definecolor{revtwocolor}{RGB}{128,0,160}
\ifrevisions
  \newcommand{\rev}[1]{\textcolor{revcolor}{#1}}
  \newcommand{\revv}[1]{\textcolor{revtwocolor}{#1}}
  \newcommand{\revon}{\color{revcolor}}
  \newcommand{\revvon}{\color{revtwocolor}}
\else
  \newcommand{\rev}[1]{#1}
  \newcommand{\revv}[1]{#1}
  \newcommand{\revon}{}
  \newcommand{\revvon}{}
\fi

\newcommand{\bw}{\boldsymbol{w}}
\newcommand{\bW}{\boldsymbol{W}}

\newcommand{\bs}{\boldsymbol{s}}

\newtheoremstyle{Astyle}
  {3pt}
  {3pt}
  {\normalfont}
  {}
  {\bfseries}
  {.}
  {0.5em}
  {\thmnumber{#2}\thmnote{. #3}}

\theoremstyle{Astyle}
\newtheorem{assumption}{Assumption}

\newenvironment{restatedassumption}[2]{%
  \par\vspace{3pt}%
  \noindent{\bfseries\ref{#1}.\ #2.}\enspace\normalfont\ignorespaces%
}{%
  \par\vspace{3pt}%
}

\theoremstyle{plain}
\newtheorem{theorem}{Theorem}[section]

\newtheorem{proposition}[theorem]{Proposition}

\theoremstyle{definition}
\newtheorem{definition}[theorem]{Definition}

\usepackage{caption}
\usepackage{makecell}

%% file: Sections/1-introduction.tex
\section{Introduction}

\label{sec:introduction}

\begin{figure}
    \centering
    \includegraphics[width=0.66\linewidth]{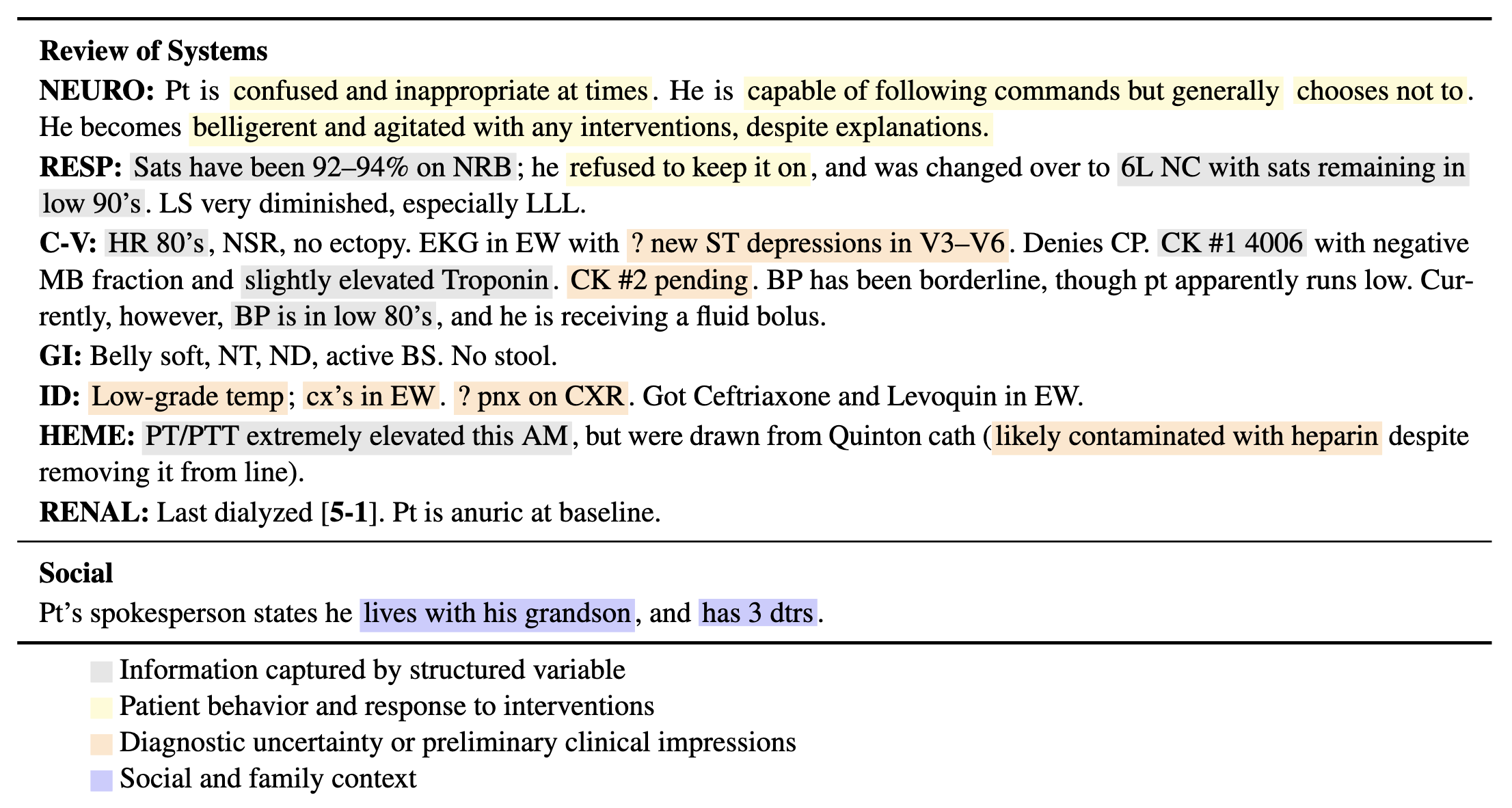}
    \caption{Example note with color-coded highlights for information often missing from structured variables. }
    \label{fig:use_case}
\end{figure}

Causal inference plays a central role across social, biomedical, and computational sciences~\citep{rubin1974estimating, holland1986statistics, Pearl-Causality}. 
A primary objective in causal inference is to estimate treatment effects, such as the average treatment effect (ATE). 
Identification of these estimands requires the assumptions of consistency, unconfoundedness, and positivity~\citep{Pearl-Causality}.
To satisfy \textit{unconfoundedness}, a common practice is to collect as much relevant information as possible. 
Beyond structured covariates, unstructured text has emerged as an additional source of information. 
Text often encodes rich contextual information that simultaneously influences treatment assignment and outcome, making it a promising source for confounding adjustment by surfacing otherwise unrecorded confounders~\citep{egami2022make, feder-etal-2022-causal, wei2022emergent, zhao2023survey}. 
A motivating example arises in critical care, where randomized trials are seldom feasible and analyses rely on observational data. 
The Medical Information Mart for Intensive Care III database~\citep[MIMIC-III,][]{johnson2016mimic} combines structured electronic health record (EHR) variables (e.g., demographics, laboratory measurements) with unstructured clinical notes. 
As illustrated in Figure~\ref{fig:use_case}, these notes capture rich contextual information about comorbidities and patient adherence, which are not encoded in structured EHRs yet may confound both treatment assignment and outcomes.

\begin{figure}[!t]
    \centering
    \includegraphics[width=0.78\linewidth]{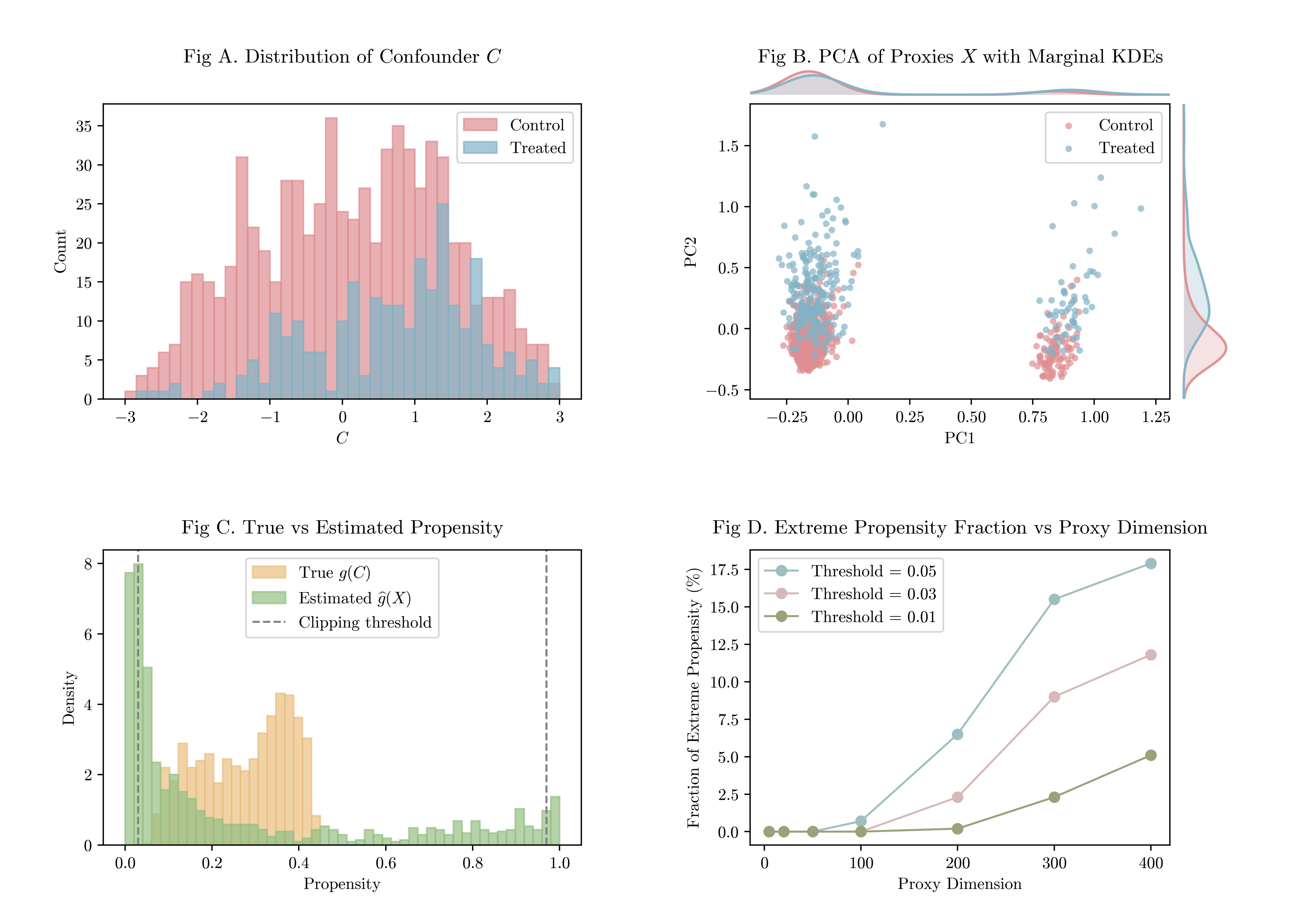}
    \caption{Illustration of observational-level positivity violations induced by high-dimensional proxy covariates. Panel A shows overlap between treated and control units in the latent confounder space. 
    Panel B visualizes the corresponding proxy space using the first two principal components and marginal KDEs. 
    Panel C compares the ground-truth and estimated propensity score distributions, with dashed lines indicating clipping thresholds. 
    Panel D shows that the clipping fraction increases with proxy dimension.}
    \label{fig:toy-example}
\end{figure}

Despite this promise, incorporating text poses unique challenges for the \textit{positivity} assumption, which requires the probabilities of treatment assignment (also known as the propensity score) to remain bounded away from 0 and 1 for all confounder combinations. 
This assumption is particularly vulnerable in text settings because (i) text often contains \textit{irrelevant or spurious features} that obscure the true confounding patterns; and (ii) the \textit{high-dimensional, sparse, and discrete} nature of text substantially enlarges the covariate space, reducing overlap between treatment groups. 
As a result, even when positivity holds in the underlying data-generating process, overlap may deteriorate in the observed text.
We refer to this mismatch as an \textit{observational-level positivity violation} (illustrated in Figure \ref{fig:toy-example}): empirical overlap fails not because positivity is violated at the population level, but owing to high-dimensional and redundant textual proxies. 
This leads to extreme estimated propensity scores near 0 or 1, producing unstable inverse probability weighted (IPW) estimates and inflated variance in estimators, as observed in prior work~\citep{belloni2014high, keller2015neural, feder-etal-2022-causal}. 
Theoretically, \citet{melnychuk2024bounds} bound confounding bias from text representations, highlighting the risks of unconstrained encoding.
These challenges motivate our focus in this paper: 
\begin{center}
    \textit{How can we extract a statistically sufficient, low-dimensional representation of text that preserves unconfoundedness while mitigating observational-level positivity violations?}
\end{center}

To address this problem, we propose selecting sparse token subsets that preserve confounding information and reduce redundant textual proxies that contribute to observational-level positivity violations. 
Our approach uses residual dependence regularization based on the Hilbert--Schmidt Independence Criterion (HSIC), a kernel-based measure widely used in independence testing and causal discovery~\citep{gretton2007kernel, zhang2011kernel, strobl2019approximate}.
The key insight is that if the selected tokens preserve
unconfoundedness, then the residuals of the treatment and outcome
models should be independent, \revv{under conditions on the outcome-generating process}.
Residual HSIC therefore provides a \textit{falsification test} \revv{for representation sufficiency, encourages retention of adjustment-relevant information and improves empirical overlap}.
Our main \textbf{contributions} are fourfold: 

$\bullet$ \textit{Conceptually}, to our knowledge, this is the first work on causal effect estimation with high-dimensional text that emphasizes interpretable text rationalization for robustness. 
We formalize how text confounding can induce \emph{observational-level positivity violations} and highlight the need for \rev{compact and confounding-relevant rationales. 
With necessity and sufficiency as its design objectives, the selector aims to identify compact and adjustment-relevant token subsets}.

$\bullet$ \textit{Methodologically}, we introduce \emph{Confounding-Aware Token Rationalization (CATR)}, a token-level rationalization framework for causal inference with text data. 
CATR produces sparse and confounding-aware text representations by minimizing the residual HSIC criterion under a sparsity penalty, \revv{encouraging} the conditional independence structure required for unconfoundedness. 
The framework provides a principled and computationally efficient approach for mitigating observational-level positivity violations and stabilizing downstream causal estimation.

$\bullet$ \textit{Theoretically}, we establish conditions under which residual HSIC provides a valid falsification test for the sufficiency of the text representation. 
\revv{We derive nonasymptotic nuisance error bounds and establish conditions for $\sqrt{n}$-consistency, asymptotic normality, and semiparametric efficiency of the augmented IPW (AIPW) estimator.}

$\bullet$ \textit{Empirically}, through semi-synthetic experiments
and a real-world analysis of the MIMIC-III sepsis cohort, we
demonstrate that CATR can improve overlap and stabilize ATE
estimation compared to existing text-based causal methods,
while providing interpretable token-level rationales.

\subsection{Related Works}
\label{subsec:related-work}

\textbf{Text-based representation learning} aims to construct low-dimensional representations of text. Early work, such as topic models~\citep{Roberts2020AdjustingFC,egami2022make} or bag-of-words~\citep{wood-doughty-etal-2018-challenges}, provides summaries of text but struggles to capture nuanced linguistic features. The advent of transformer-based language models, such as BERT~\citep{devlin2019bertpretrainingdeepbidirectional}, marked a shift toward context-aware text embeddings. 
Building on these advances, two lines of work further explore representation learning for causal inference with text.
The first learns task-specific embeddings: CausalBERT~\citep{pmlr-v124-veitch20a} jointly predicts treatment and outcome, while \citet{gui2023causal} address overlap when treatment is a text attribute.
The second line employs latent-variable models, including
CEVAE~\citep{louizos2017causal}, TEDVAE~\citep{zhang2021treatment},
and DIVA~\citep{zhou2023causal}, to infer latent confounders
and disentangle adjustment-relevant factors.
While these methods reduce dimensionality, they do \textit{not directly address the observational-level positivity violation} caused by redundant and high-dimensional textual proxies. 
Moreover, latent factors can be \textit{difficult to interpret and sensitive} to model assumptions.

\textbf{Propensity score stabilization} is commonly used to mitigate limited overlap and extreme inverse probability weights in settings with structured covariates.
One class of methods trims observations with extreme propensity scores or clips the corresponding weights~\citep{crump2009dealing,lee2011weight}, while others rescale weights to control variance inflation~\citep{cole2008constructing, li2018balancing, van2025stabilized}. 
Another class, covariate balancing~\citep{imai2014covariate}, directly enforces covariate balance between treatment groups during estimation.
However, these techniques are \textit{not well suited to textual settings}: 
post-hoc adjustments, such as trimming, discard valid samples without resolving covariate imbalance; balancing methods lack explicit semantic dimensions over which covariate balance can be enforced.

\textbf{Selective rationalization} methods aim to extract concise token subsets that preserve predictive performance~\citep{bao-etal-2018-deriving, jain-etal-2020-learning, guerreiro-martins-2021-spectra, antognini2021rationalizationconcepts}.
Early approaches used non-differentiable selectors~\citep{lei-etal-2016-rationalizing, yu-etal-2019-rethinking}, while later work introduced differentiable relaxations such as the Gumbel–Softmax for end-to-end training~\citep{bastings-etal-2019-interpretable, geng-etal-2020-selective, Sha_Camburu_Lukasiewicz_2021, liu2022frfoldedrationalizationunified}. 
Recent efforts have connected rationalization to causal notions of necessity and sufficiency, though predominantly in \textit{predictive} rather than inference settings~\citep{shiyu2020invariant,cai2023learning,zhang2023towards}.

%% file: Sections/2-notation.tex
\section{Problem Setup}
\label{sec:notation}

\subsection{Notations}

\begin{wrapfigure}{r}{0.36\textwidth}
    \vspace{-15pt}  
    \centering
    \includegraphics[width=0.7\linewidth]{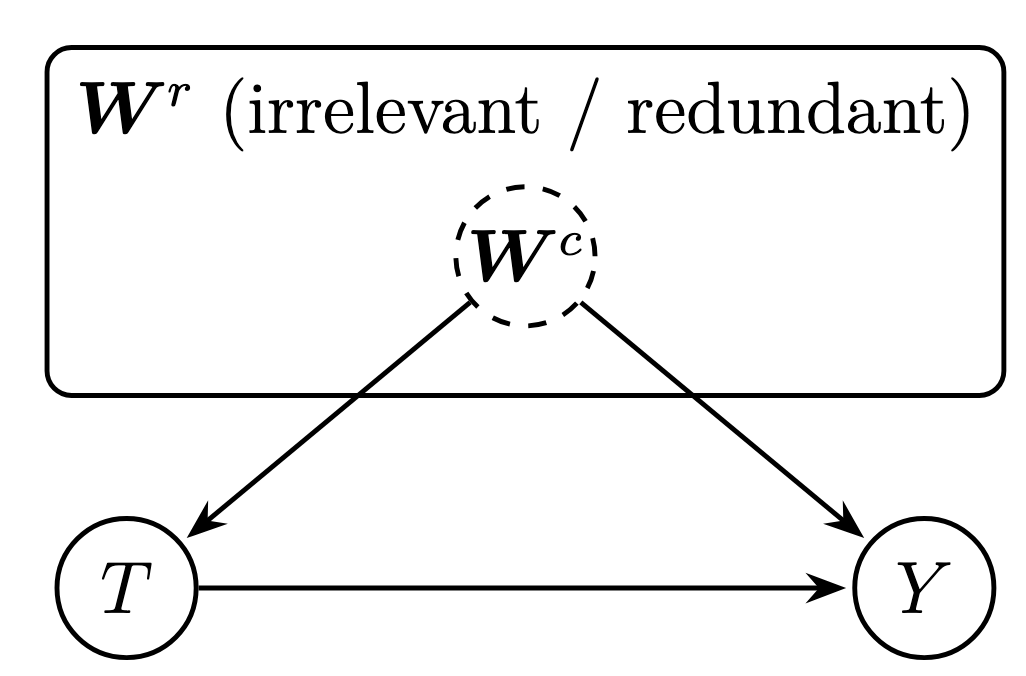}
    \caption{Causal diagram illustrating text-based confounding:
    $\bW^c$ affects both $T$ and $Y$, while $\bW^r$ denotes irrelevant or redundant tokens that the causal pathways do not operate through.}
    \vspace{-15pt}  
    \label{fig:causal}
\end{wrapfigure}

Let $(\bW,T,Y)$ denote random variables for the text, treatment, and outcome, respectively. 
We observe $n$ independent and identically distributed samples $\{(\bw_i, t_i, y_i)\}_{i=1}^n,$ where $\bw_i=(w_{i1},\ldots,w_{iL_i})$ is the observed text for unit $i$ of length $L_i$, $t_i\in\{0,1\}$ is the binary treatment indicator, and $y_i\in\{0,1\}$ (or $\mathbb R$) is the observed outcome. Let \( \Lambda(\boldsymbol{W}) \) denote the embedding of $\boldsymbol{W}$ produced by a pre-trained encoder $\Lambda(\bullet)$. Following the potential outcome framework~\citep{rubin1974estimating}, each unit has two potential outcomes, $Y(0)$ and $Y(1)$, under control and treatment, respectively.  

As discussed in Section~\ref{sec:introduction}, although the full text $\bW$ contains rich contextual information, only a subset $\bW^c\subset \bW$ encodes the true confounding signals that jointly affect both $T$ and $Y$. 
The remaining tokens, denoted as $\bW^r$, are irrelevant or redundant. 
Throughout, we adopt the causal structure illustrated in Figure~\ref{fig:causal}. 

We define the true propensity score and mean outcome functions as $g({\bw^c}) = \Pr(T\!=\!1 \!\mid\!{\bW^c\!=\!\bw^c})$ and $Q_t({\bw^c}) = \E[Y\!\mid\! T\!=\!t, {\bW^c\!=\!\bw^c}]$, respectively. 
Let $\widehat g(\cdot)$ and $\widehat Q_t(\cdot)$ denote their corresponding estimators, which may be constructed using either the full text $\bW$ or a selected subset.
We are interested in the estimation of the average treatment effect (ATE), defined as $\tau = \E\{Y(1)-Y(0)\}$. 
Its identification from observational data requires the following assumptions:
 
\begin{assumption}[Unconfoundedness]
\label{assumption:unconfoundedness}
    $\bW^c$ are sufficient for confounding adjustment:
        \(\{Y(0), Y(1)\} \perp T \mid \bW^c.\) 
\end{assumption}

\begin{assumption}[Positivity on True Signals]
\label{assumption:positivity}
    Each unit has a nonzero probability of receiving both treatment and control conditioned on $\bW^c$:
    \(0 < p_1 \leq \Pr(T=1 \mid \bW^c)\leq p_2 < 1, \quad \forall  \bW^c.\)
\end{assumption}

\begin{assumption}[Consistency]
\label{assumption:consistency}
    The observed outcome corresponds to the potential outcome under the received treatment: $Y=Y(T)$.
\end{assumption}

Under Assumptions~\ref{assumption:unconfoundedness}-- \ref{assumption:consistency}, the average treatment effect $\tau$ is identifiable. 
Given an estimated propensity score $\widehat g$ and estimated conditional outcomes $\{\widehat Q_t\}_{t\in\{0,1\}}$, we consider common estimators for $\tau$, including outcome regression \citep[OR,][]{Pearl-Causality}, inverse propensity weighting \citep[IPW,][]{rosenbaum1983central}, and augmented inverse propensity weighting \citep[AIPW,][]{bang2005doubly}:
\begingroup
\small
\setlength{\jot}{1pt}                % 公式行间距
\setlength{\abovedisplayskip}{10pt}  % 与上方正文的距离
\setlength{\belowdisplayskip}{10pt}  % 与下方正文的距离
\setlength{\abovedisplayshortskip}{10pt}
\setlength{\belowdisplayshortskip}{10pt}
\begin{align}
\widehat\tau_{\mathrm{OR}}
&= \frac{1}{n}\sum\nolimits_{i=1}^n
\left[\widehat Q_{1,i}-\widehat Q_{0,i}\right],
\label{eq:or}\\
\widehat\tau_{\mathrm{IPW}}
&= \frac{1}{n}\sum\nolimits_{i=1}^n
\left[
\frac{t_i y_i}{\widehat g_i}
-\frac{(1-t_i)y_i}{1-\widehat g_i}
\right],
\label{eq:ipw}\\
\widehat\tau_{\mathrm{AIPW}}
&= \frac{1}{n}\sum\nolimits_{i=1}^n
\left[
\frac{t_i(y_i-\widehat Q_{1,i})}{\widehat g_i}
+\widehat Q_{1,i}
-\frac{(1-t_i)(y_i-\widehat Q_{0,i})}{1-\widehat g_i}
-\widehat Q_{0,i}
\right].
\label{eq:aipw}
\end{align}
\endgroup
where $\widehat g_i$ and $\widehat Q_{t,i}$ denote the estimated propensity score and conditional outcome for unit $i$.  
While these assumptions are standard for structured covariates, extending them to text introduces additional complexities.
In particular, there is a tension between unconfoundedness and positivity: incorporating rich text can help account for confounding, while its high dimensionality can weaken positivity.

\subsection{Observational-level Positivity Violation}\label{sec:vio}

Since the true confounders $\bW^c$ are unknown, analysts often condition on the full text $\bW$, which leads to high-dimensional covariates, diminished empirical overlap, and unstable propensity score estimates.  

We illustrate this effect through a controlled simulation, where treatment assignment depends on a single latent confounder \(C\) and positivity holds by construction. 
However, the observed covariates \(X\) are high-dimensional sparse proxies for \(C\), mirroring the empirical behavior of high-dimensional text confounders. 
As shown in Figure~\ref{fig:toy-example}, treated and control units overlap well in the latent \(C\)-space, but their separation becomes more pronounced after mapping to the proxy \(X\)-space. 
This distortion leads the estimated propensities \(\widehat g(X)\) to place more mass near 0 and 1 than the ground-truth propensities, often crossing clipping thresholds at 0.03 and 0.97. 
The resulting inverse probability weights become heavy-tailed. 
As the proxy dimension increases, the clipping fraction also rises.

%More specifically, because $\bW^c$ cannot be directly separated from $\bW$, analysts often condition on the full text $\bW$, leading to high-dimensional covariates, unstable propensity score estimates, and limited overlap between treatment groups. 

We refer to this phenomenon as an \textit{observational-level positivity violation}: although the true treatment mechanism \(g(\bW^c)\) is well balanced, the estimated propensity scores \(\widehat g(\bW)\) can become unstable when redundant or irrelevant tokens distort the empirical covariate space. 
In this case, effective overlap deteriorates even though Assumption~\ref{assumption:positivity} holds at the latent confounder level.
Formally, we define:

\begin{definition}[Observational-level positivity violation]
\label{def:violation}
Let $g(\bw)\!=\!\Pr(T=1\!\mid\!\bW\!=\!\bw)$ be the treatment mechanism, and $\widehat g(\bw)\!=\!\Pr_n(T\!=\!1\!\mid\!\bW\!=\!\bw)$ its empirical estimate. 
An \emph{observational-level positivity violation} occurs when there exists $\bw \in \mathrm{supp}(\bW)$ such that $\widehat g(\bw) \notin [\epsilon, 1-\epsilon]$ for some $0 < \epsilon < \frac12$, even though the true treatment assignment mechanism satisfies $g(\bw^c) \in [\epsilon, 1-\epsilon]$, given the true confounding subset $\bw^c$. 
\end{definition}

Importantly, this notion differs from a positivity violation in the underlying data-generating process, which leads to non-identifiable causal effects. 
Observational-level violations, instead, preserve identifiability but complicate estimation by degrading overlap and producing unstable inverse-propensity weights.

%% file: Sections/3-method.tex
\section{Methodology}
\label{sec:method}
In this section, we formally introduce Confounding-Aware Token Rationalization (CATR), a framework that integrates HSIC-based residual independence into token-level rationalization, as illustrated in Figure~\ref{fig:text_tabular_fusion} and summarized in Algorithm~\ref{alg:train}.

\subsection{Text Selection}
\label{subsec:selection}
 
To mitigate the observational-level positivity violation in Definition \ref{def:violation}, our goal is to reduce the irrelevant or redundant information by learning a text selector \[S(\cdot):\bw_i \longrightarrow  \widetilde{\bw}_i=S(\bw_i),\] which maps each text sequence $\bw_i$ to \rev{an adjusted representation} $\widetilde{\bw}_i$ \rev{concentrated on an effective number of tokens} $d_i \ll L_i$, \revv{with the aim of retaining the adjustment-relevant information in} $\bw_i^c$. 
\rev{We write $\widetilde{\bW}=S_\theta(\bW)$ throughout.} 
To ensure interpretability and statistical stability, the selector is subject to a sparsity constraint, such as a token budget or an entropy-based regularization. 
\rev{For the identification of $\tau$ to transfer from $\bW^c$ to the learned representation, $\widetilde{\bW}$ must satisfy the following conditions:}
\[
\underbrace{(Y(0),Y(1)) \perp T \mid \widetilde{\bW}}_{\text{Unconfoundedness}}, \qquad
\underbrace{\epsilon \le g(\widetilde{\bW}) \le 1-\epsilon}_{\text{Positivity}}, \qquad
\underbrace{Y = Y(T)}_{\text{Consistency}}.
\]
\rev{While consistency holds by construction if the treatment and outcome are well-defined, the remaining conditions cannot be imposed directly: unconfoundedness involves potential outcomes that are never jointly observed, and positivity is a property of the assignment mechanism whose direct optimization would trade against adjustment-relevant information.
This motivates our objective (Equation~\ref{eq:loss}) that combines a testable
implication of unconfoundedness and a sparsity criterion as
attainable proxies.}

\subsection{HSIC Measurement for Subset Sufficiency}
\label{sec:hsic}
To navigate the trade-off between improving empirical overlap and retaining confounding information, we introduce a penalty based on the Hilbert–Schmidt Independence Criterion~\citep[HSIC,][]{gretton2007kernel}. 
The key idea is that, conditioning on a sufficient text representation, the residuals of the treatment and outcome models should be independent. % HSIC provides an efficient and fully nonparametric diagnostic for assessing this residual dependence during rationalization.

\begin{definition}[Hilbert--Schmidt Independence Criterion]
\label{def:hsic}
Given two random variables $U$ and $V$, let $k_U$ and $k_V$
be characteristic kernels with Gram matrices
$K,L\in\mathbb{R}^{n\times n}$ over $n$ samples.
The empirical HSIC is
\[
\widehat{\mathrm{HSIC}}(U,V)
= \frac{1}{(n-1)^2}\mathrm{tr}(KHLH),
\]
where $H=I_n-\frac{1}{n}\mathbf{1}_n\mathbf{1}_n^\top$,
$I_n$ is the identity matrix, $\mathbf{1}_n$ is the vector
of ones, and $\mathrm{tr}(\cdot)$ denotes the matrix trace.
\revv{The corresponding population} HSIC is nonnegative
and equals zero if and only if $U\perp V$.
\end{definition}

\revv{To establish the one-sided diagnostic role of residual HSIC in assessing representation sufficiency}, we impose the following outcome-noise condition~\citep{Hoyer2009,Peters2014causal,strobl2019approximate}. Formally, we assume:

\begin{assumption}[\rev{Noise Structure}]
\label{assumption:noise}

There exist measurable functions $g^\star:\mathcal{W}^c \to [0,1]$ and $Q_T^\star:\{0,1\}\times\mathcal{W}^c \to \mathbb{R}$ such that
\rev{
\(
T \mid \bW^c \sim \mathrm{Bernoulli}\bigl(g^\star(\bW^c)\bigr),\ 
Y = Q_T^\star(\bW^c) + \varepsilon_Y,
\)
and the outcome noise $\varepsilon_Y$ satisfies
(i) $\varepsilon_Y \perp (T,\bW)$ and
(ii) $\E[\varepsilon_Y] = 0$, $\E[\varepsilon_Y^2] < \infty$.}
\end{assumption}

\rev{The Bernoulli model accommodates binary treatment.
The outcome-noise condition is most appropriate for continuous
outcomes with approximately independent additive noise;
heteroskedasticity, treatment-dependent noise, or discrete
outcomes may require more flexible assumptions.}

Under Assumption~\ref{assumption:noise}, if $\widetilde{\bW}$ \revv{satisfies the sufficiency conditions}, the residual dependence between treatment and outcome vanishes, as detailed in the following proposition. Proof is provided in Appendix~\ref{sec:sproof}.

\begin{proposition}
\label{prop:hsic}
Let $\widetilde{\bW}\subseteq\bW$ be any candidate subset. Define $g(\widetilde{\bW})= \E[T\mid\widetilde{\bW}]$ and $Q_t(\widetilde{\bW}) = \E[Y\mid T=t,\,\widetilde{\bW}]$, as well as residuals $r_T = T - g(\widetilde{\bW})$ and $r_Y = Y - Q_T(\widetilde{\bW})$.  
Let $D(\widetilde{\bW}) = \mathrm{HSIC}(r_T,r_Y)$ be the population-level HSIC computed with characteristic and bounded kernels under finite second moments.
Suppose Assumptions~\ref{assumption:unconfoundedness},~\ref{assumption:consistency}, and~\ref{assumption:noise} hold. 
\rev{$\widetilde{\bW}$ is \emph{sufficient} if
\begin{enumerate}[label=(\roman*),itemsep=0pt,topsep=2pt,parsep=0pt,partopsep=0pt]
\item $(Y(0),Y(1)) \perp T \mid \widetilde{\bW}$, and
\item $Q_t^\star(\bW^c) = \widetilde Q_t(\widetilde{\bW})$ for some $\widetilde Q_t$, $t\in\{0,1\}$.
\end{enumerate}}
Then,
\[
\text{$\widetilde{\bW}$ \rev{is sufficient}}
\quad\Longrightarrow\quad
D(\widetilde{\bW}) = 0.
\]
Contrapositively,
\[ D(\widetilde{\bW}) > 0 \quad\Longrightarrow\quad \text{\rev{$\widetilde{\bW}$ is not sufficient}}. \]
\end{proposition}

\revv{Here, insufficiency means failure of at least one of conditions (i) and (ii). The supervised outcome loss encourages $\widetilde{\bW}$ to retain outcome-predictive information, targeting (ii).
Intuitively, penalizing residual dependence further encourages retention of adjustment-relevant information, supporting unconfoundedness in (i). Residual HSIC thus provides a falsification diagnostic for the joint sufficiency conditions.}

This method follows HSIC-based independence testing~\citep{gretton2005measuring,gretton2007kernel} and its application in feature selection and causal discovery~\citep{song2012feature,zhang2011kernel,strobl2019approximate}. 
Following the logic of falsification checks in observational studies~\citep[e.g.,][]{pearl1995testability, karlsson2025falsification}, \revv{we use this one-sided diagnostic to guide representation learning.} 
We incorporate it as a regularization term in the training objective (Equation~\ref{eq:loss}) \revv{to encourage low residual dependence alongside nuisance prediction and sparsity.}

\subsection{Architecture and Training Objective}
\label{subsec:objective}
Building upon the standard rationalization paradigm~\citep[see e.g.,][]{jain-etal-2020-learning,antognini2021rationalizationconcepts}, our framework consists of a selector $S(\cdot)$ and predictors $g(\cdot)$, $\{Q_t(\cdot)\}_{t\in\{0,1\}}$.
For each observed text $\bw_i$, the selector $S_\theta(\cdot)$ produces a token-level mask $s_i=S_\theta(\bw_i)\in [0,1]^{L_i}$ parameterized by $\theta$.
During training, $s_i$ is drawn from a relaxed Bernoulli distribution to introduce stochasticity and encourage exploration of alternative subsets. 
During evaluation, we instead conduct deterministic selection \rev{by using the continuous scores directly as pooling weights, which can be viewed as a form of soft selection. 
Appendix~\ref{app:hardselect} examines several hard-selection rules and finds the estimates robust to this choice, since the learned weights are already concentrated on a small number of tokens.}

The resulting weighted token representation is $\widetilde \bw_i\!=\!s_i \odot \bw_i,$ where $\odot$ denotes elementwise multiplication. Given $\widetilde \bw_i$,
the predictor network, parameterized by $\phi$, produces the estimated propensity score $\widehat g(\widetilde\bw_i)$ and conditional outcomes $\widehat Q_t(\widetilde\bw_i)$.
The architecture conceptually parallels DragonNet~\citep{shi2019adapting}, except that the representation $\widetilde\bw$ is itself learnable through the selector $S_\theta(\cdot)$. 
This design enables token-level control of confounding adjustment and provides interpretable rationales of which linguistic features contribute most.

\begin{figure}[!t]
    \centering
    \includegraphics[width=0.75\textwidth]{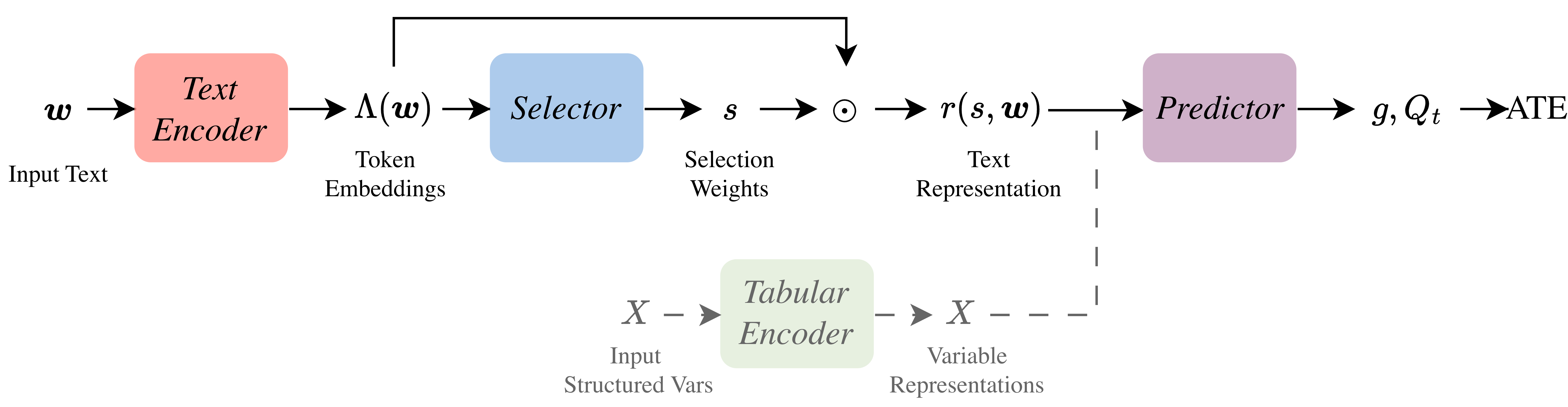}
    \caption{The CATR framework (with multimodal extension). The main pipeline (solid blocks) implements selective rationalization for text-only inputs. The semi-transparent dashed components illustrate the extension to tabular covariates, showing that our approach naturally adapts to multi-modal settings.}
    \label{fig:text_tabular_fusion}
\end{figure}

Given a training dataset $\mathcal D=\{(\bw_i, t_i, y_i)\}_{i=1}^n$, we formalize this architecture through the following joint training objective:
\begin{equation}
    \min_{\theta,\phi}\; 
\frac{1}{n}\sum_{i=1}^n 
\Big[
\mathcal{L}_{\text{sup}, i} 
+ \mu\,\mathcal{R}_{\text{sparse}}(s_i)
+ \gamma\,\mathrm{HSIC}(r_{T,i},r_{Y,i})
\Big],
\label{eq:loss}
\end{equation} 
where $s_i=S_\theta(\bw_i)$ is the selection mask, and $\mu,\gamma>0$ control the trade-off among predictive accuracy, sparsity, and residual independence. 

The loss consists of three components:

\noindent\textbf{(i) Supervised loss.} $\mathcal{L}_{\text{sup},i}$ trains the propensity score and potential outcome predictors, ensuring predictive fidelity. Formally, 
\[
\mathcal{L}_{\text{sup},i}
= {\revon \alpha\,}\mathrm{CE}\big(t_i, \widehat g_\phi(\widetilde \bw_i)\big) 
+ \beta\,\mathrm{CE}\big(y_i, \widehat Q_{t_i,\phi}(\widetilde \bw_i)\big),
\]
where $\mathrm{CE}(\cdot,\cdot)$ is the cross-entropy loss and \rev{$\alpha,\beta>0$ weight} the supervised losses.

\noindent\textbf{(ii) Sparsity regularization.} 
$\mathcal{R}_{\text{sparse}}(s_i)$ is a general regularization term that encourages compact token subsets. 
Depending on the desired sparsity level and behavior, the regularizer can take different forms: a Kullback--Leibler divergence term with a $\mathrm{Bernoulli}(\pi)$ prior to enforce a token budget $\pi$; an entropy regularization to promote confident selections; or norm-based penalties for direct sparsity control. 
In all experiments, we use entropy regularization to avoid additional hyperparameter tuning.

\noindent\textbf{(iii) HSIC penalty.}
$\mathrm{HSIC}(r_{T,i},r_{Y,i})$ measures residual dependence between treatment and outcome. 
In practice, this term is evaluated as a batch-level statistic $\widehat{\mathrm{HSIC}}(\{r_{T,i}\}_{i=1}^m, \{r_{Y,i}\}_{i=1}^m)$ over the current minibatch. We use per-sample notation in Equation~\ref{eq:loss} for simplification.
We compute HSIC using characteristic kernels: a linear kernel $k_T(u,v)=u^\top v$ for the binary treatment residual $r_T$, and a Gaussian RBF kernel $k_Y(u,v)=\exp(-\|u-v\|^2/2\sigma_Y^2)$ for the continuous outcome residual $r_Y$. 
The bandwidth $\sigma_Y$ is selected using the median heuristic. 
This choice follows common practice in independence testing. 

As stated in Proposition~\ref{prop:hsic}, \revv{HSIC provides a one-sided diagnostic: positive HSIC indicates failure of the sufficiency conditions, and minimizing empirical HSIC encourages low residual dependence and retention of adjustment-relevant information.}

In summary, the objective shapes the representation on which the nuisance functions are estimated: the supervised loss ensures predictive fidelity, the sparsity regularizer enforces concise token subsets under user-defined constraints, and the HSIC term promotes residual independence, thereby stabilizing causal effect estimation. 

At inference time, the trained $\widehat g(\widetilde\bw_i)$ and $\{\widehat Q_t(\widetilde\bw_i)\}$ are substituted into Equations~(\ref{eq:or})--(\ref{eq:aipw}) as plug-in estimates, yielding $\widehat\tau_{\mathrm{OR}}$, $\widehat\tau_{\mathrm{IPW}}$, and $\widehat\tau_{\mathrm{AIPW}}$. 
The hyperparameter selection and implementation details are provided in Appendix~\ref{sec:implementation}. \rev{Appendix~\ref{app:sensitivity} reports sensitivity to $\mu$ and $\gamma$.}

\begin{algorithm}[!t]
\footnotesize
\caption{Confounding-Aware Token Rationalization (CATR)}
\label{alg:train}
\let\AND\relax
\begin{algorithmic}[1]
   \STATE \textbf{Input:} dataset $\mathcal D=\{(\bw_i, t_i, y_i)\}_{i=1}^n$; temperature $\eta$; prior budget $\pi$; loss weights $(\mu,\gamma)$
   \STATE Split $\mathcal D$ into training and validation sets
   \STATE Initialize selector $S(\cdot)$, predictor $g(\cdot)$, and $\{Q_{t}(\cdot)\}_{t\in\{0,1\}}$
   \WHILE{not converged}
       \STATE Sample minibatch $\{(\bw_i, t_i, y_i)\}_{i=1}^m \sim \mathcal D_{\mathrm{train}}$
       \STATE Compute selection score $a_{i}=S_\theta(\bw_{i})$
       \STATE Sample relaxed gates $s_{ij}\sim\mathrm{RelaxedBernoulli}(\eta,a_{i})$ to form a differentiable binary mask  
       \STATE Apply mask to obtain $\widetilde\bw_i = s_i \odot \bw_i$
       \STATE Predictor outputs estimated propensity score $\widehat g(\widetilde\bw_i)$ 
       and conditional mean outcomes $\{\widehat Q_{t}(\widetilde\bw_i)\}_{t\in\{0,1\}}$
       \STATE Form residuals $r_{T,i}=t_i-\widehat g_i$, $r_{Y,i}=y_i-\widehat Q_{t_i,i}$
       \STATE Compute loss in Equation~\ref{eq:loss}
       \STATE Update model parameters via stochastic gradient descent
   \ENDWHILE
   \STATE \textbf{Output:} trained selector $S(\cdot)$, predictors $g(\cdot)$ and $\{Q_{t}(\cdot)\}_{t\in\{0,1\}}$
\end{algorithmic}
\end{algorithm}

\subsection{Extension: Multimodal Data}
\label{subsec:extension}

While our main method is developed for text-only settings, many applications also include well-curated structured covariates. 
In such cases, text serves as a complementary source of confounding information. 
We extend our framework to jointly handle textual and structured data by encoding tabular covariates into dense vectors of the same dimension as text embeddings, and fusing them into a single vector, following the strategy of \citet{ebrahimi2023lanistr}. 
As shown in Figure~\ref{fig:text_tabular_fusion}, the token selection mechanism continues to operate on the text encoder output in this end-to-end differentiable architecture, while the fused representation allows joint adjustment for text and structured covariates without modifying the core architecture.

% We view this as an extension to our method rather than a primary methodological contribution. Nevertheless, developing more efficient integration strategies for structured and unstructured data in causal inference remains a promising future direction.

%% file: Sections/4-theoretical.tex
\section{Theoretical Analysis}
\label{sec:theory}

This section establishes the statistical properties of the proposed CATR framework. 
All proofs are provided in Appendix~\ref{sec:sproof}.

\subsection{Nonasymptotic Error Bound of Neural Network Approximation}

We begin by introducing a set of regularity conditions required for our theoretical results.

\begin{assumption}[Regularity of data and representation]
\label{assumption:data}
The treatment $T\in\{0,1\}$ and outcome $Y\in[-M,M]$ for some $M>0$. 
The observed text $\bW$ is encoded by a pretrained model into a bounded, continuous-valued feature vector $\mathbf{x}=\Lambda(\bW)\in[-1,1]^d$, whose distribution admits a bounded density on $[-1,1]^d$, where $d$ is the encoder output dimension.
\end{assumption}

\begin{assumption}[Smoothness]
\label{assumption:smoothness}
For each target function $f\in\{g,Q_0,Q_1\}$, assume that $f$ lies in a Sobolev ball $\mathcal{W}^{\beta,\infty}([-1,1]^d)$ with smoothness $\beta\in\mathbb{N}_{+}$, i.e.,
\[
f(\mathbf{x})\in\mathcal{W}^{\beta,\infty}([-1,1]^d)
:=\Bigl\{
f:\max_{|\alpha|\le\beta}\mathrm{ess\,sup}_{\mathbf{x}\in[-1,1]^d}
|D^{\alpha}f(\mathbf{x})|\le 1
\Bigr\},
\]
where $\alpha=(\alpha_1,\ldots,\alpha_d)$, $|\alpha|=\sum_j \alpha_j$, and $D^{\alpha}f$ denotes the weak derivative.
\end{assumption}

% \begin{assumption}[Regularity of penalization terms]
% \label{assumption:penalty}
% The regularization terms $\mathcal{R}_{\text{sparse}}$ and $\mathrm{HSIC}$ 
% are nonnegative and uniformly bounded.
% \end{assumption}

These assumptions are mild in language-based settings: pretrained encoders yield bounded, smoothly varying embeddings, where small semantic changes lead to gradual shifts in propensity or conditional outcome. Under the above assumptions, we establish the nonasymptotic error bound for deep MLP-ReLU network approximation following the framework of~\citet{Farrell_2021}.

\begin{theorem}
\label{thm:penalized}
Suppose Assumptions~\ref{assumption:data}--\ref{assumption:smoothness} hold.
Let $\widehat f$ be the minimizer of~\eqref{eq:loss}, \rev{with penalty weights $\mu_n,\gamma_n>0$.
Suppose $\mathcal{R}_{\mathrm{sparse}}$ and $\mathrm{HSIC}$ are uniformly bounded by $C_{\mathrm{sp}}$ and $C_{\mathrm{HSIC}}$, and write}
{\revon
\(
a_n = n^{-\frac{\beta}{\beta+d}}\log^8 n + \frac{\log\log n}{n},~
b_n = \mu_n C_{\mathrm{sp}} + \gamma_n C_{\mathrm{HSIC}}.
\)}
\revv{Suppose the nuisance functions and the selector are jointly estimated over an MLP-ReLU network class with width} $H\asymp n^{\frac{d}{2(\beta+d)}}\log^2 n$ \revv{and depth $D\asymp\log n$. 
Then} there exists $C>0$ such that, with probability at least $1-\exp(-n^{\frac{d}{\beta+d}}\log^8 n)$,
\rev{\[
\|\widehat f - f_*\|_{L_2(P_X)}^2
\le
C\bigl(a_n + b_n\bigr),
\]}%
where $f$ represents either $g$ or $Q_t$, and $f_*$ denotes the true target function. %, $d$ is the dimensionality of the text embeddings, $n$ is the sample size, $\beta$ is the Hölder smoothness parameter of the target function.
\end{theorem}
Theorem~\ref{thm:penalized} extends the standard consistency result in~\citet{Farrell_2021} to the case with additional regularization. 
\rev{Neither the HSIC term nor the sparsity term alters the statistical order of estimation error once $b_n=O(a_n)$.}
\revv{This decay condition specifies the tuning sequences required for the asymptotic results. In experiments, we use fixed penalty weights and examine sensitivity to these choices in Appendix~\ref{app:sensitivity}.}
 % Hence, reducing the dimensionality of text by selecting smaller subsets of tokens can improve the propensity estimation accuracy.

Furthermore, while $d$ corresponds to the embedding size, selective rationalization reduces the effective input complexity by removing spurious or irrelevant components and improves convergence in practice.

\subsection{Statistical Inference}

We first show the consistency and convergence of nuisance estimators. 

\begin{theorem}
\label{thm:consistency}
Suppose Assumptions~\ref{assumption:unconfoundedness}--\ref{assumption:smoothness} hold, and the nonasymptotic bound in Theorem~\ref{thm:penalized} holds \rev{with $b_n=o(1)$ for part (i) and $a_n+b_n=o(n^{-1/2})$ for parts (ii) and (iii)}. Then the estimators $\widehat g$ and $\widehat Q_t$ satisfy, for $t\in\{0,1\}$:

\begin{enumerate}[label=(\roman*),itemsep=0pt,topsep=0pt,parsep=0pt,partopsep=0pt]
\item $\E_n[(\widehat g(\bw_i)-g(\bw_i))^2] = o_P(1)$, \quad
      $\E_n[(\widehat Q_t(\bw_i)-Q_t(\bw_i))^2] = o_P(1)$;

\item $\E_n[(\widehat Q_t(\bw_i)-Q_t(\bw_i))^2]
        \E_n[(\widehat g(\bw_i)-g(\bw_i))^2]
        = o_P(n^{-1})$;

\item $\E_n[(\widehat Q_t(\bw_i)-Q_t(\bw_i))(t_i-g(\bw_i))]
        = o_P(n^{-1/2})$.
\end{enumerate}

Let $\widehat\psi_t$ denote the plug-in orthogonal score, defined as
\(
\widehat\psi_1
=\frac{t_i}{\hat g(\bw_i)}\{y_i-\widehat Q_1(\bw_i)\}
+\widehat Q_1(\bw_i),~
\widehat\psi_0
=\frac{1-t_i}{1-\hat g(\bw_i)}\{y_i-\widehat Q_0(\bw_i)\}
+\widehat Q_0(\bw_i).
\)
If $\widehat g(\bw_i)$ is bounded away from $0$ and $1$, then the empirical mean and second moment of the estimated score converge to their population counterparts:
\[
\sqrt{n}\,\E_n[\widehat\psi_t-\psi_t]
= o_P(1),
\quad
\E_n[(\widehat\psi_t)^2]-\E_n[(\psi_t)^2]=o_P(1).
\]
\end{theorem}
Theorem~\ref{thm:consistency} emphasizes the importance of observational-level positivity for the validity of nuisance function estimation. 
When it is violated, the inverse weighting terms $1/\widehat g(\bw)$ and $1/\{1-\widehat g(\bw)\}$ may diverge, causing the orthogonal scores to have unbounded variance. 
Positivity helps control both sampling fluctuations and the remaining errors from nuisance estimation in Theorem~\ref{thm:consistency}, thereby supporting finite-variance estimation of treatment effects.

Building on these properties, we analyze the convergence and asymptotic normality of the ATE estimators. 

\begin{theorem}
\label{thm:ipw-aipw}
Suppose Assumptions~\ref{assumption:unconfoundedness}--\ref{assumption:smoothness} hold.  
Define $\widehat\tau_{\mathrm{IPW}}$ and $\widehat\tau_{\mathrm{AIPW}}$ as in Equations~(\ref{eq:ipw}) and~(\ref{eq:aipw}). Let $\tau$ denote the ground truth ATE.

\begin{enumerate}[label=(\roman*),itemsep=0pt,topsep=0pt,parsep=0pt,partopsep=0pt]
\item \emph{(Consistency)}
\revv{If the nuisance consistency conditions in
Theorem~\ref{thm:consistency}(i) hold and
$\widehat g$ is bounded away from $0$ and $1$, then}
$\widehat\tau_{\mathrm{IPW}},\widehat\tau_{\mathrm{AIPW}}
\overset{P}{\to} \tau$.
\item \emph{(AIPW asymptotic normality)}
\revv{If, additionally, the remaining conditions
of Theorem~\ref{thm:consistency} hold, together with the
nuisance product-rate condition}
\(\|\widehat g-g\|_{L_2(P_{\bw})}
\cdot
\|\widehat Q_t-Q_t\|_{L_2(P_{\bw})}
=o_P(n^{-1/2}),\)
then 
\[
\sqrt{n}\,(\widehat\tau_{\mathrm{AIPW}}-\tau)
\ \rightsquigarrow\ 
\mathcal{N}(0,\,V^*),
\quad
V^*=\Var\big(\psi_1-\psi_0\big).
\]
\end{enumerate}
\end{theorem}

\revv{By Theorem~\ref{thm:penalized}, the required rate follows when $a_n\!+\!b_n\!=\!o(n^{-1/2})$.
This holds if both $a_n,b_n\!=\!o(n^{-1/2})$. $a_n\!=\!o(n^{-1/2})$ when $\beta\!>\!d$, which requires strong smoothness in
high-dimensional settings, while $b_n\!=\!o(n^{-1/2})$
requires regularization weights to decay sufficiently fast that the total penalty contribution is negligible.}

Theorem~\ref{thm:ipw-aipw} shows that while both IPW and AIPW estimators are consistent under overlap, only AIPW achieves $\sqrt{n}$-asymptotic normality under the weaker product-rate condition, which aligns with our experimental results in Section~\ref{sec:experiments}.

%% file: Sections/5-experiment.tex
\section{Empirical Experiments}
\label{sec:experiments}

In this section, we evaluate the proposed CATR  using both controlled semi-synthetic studies and a real-world analysis of the MIMIC-III sepsis cohort.

\subsection{Semi-Synthetic Experiment}

\textbf{Semi-Synthetic Data Generation.} We construct semi-synthetic data with known treatment effects, using the MIMIC-III database~\citep{johnson2016mimic}. 
We focus on a subset of ICU patients with sepsis. 
For each patient $i$, we extract the earliest physician note within eight hours of admission as $\bw_i$. 
To allow for benchmarking of causal estimators under a realistic but controlled setting, we simulate treatment assignments and outcomes directly from the clinical notes. 
We first construct binary indicators based on the presence of some infection-related keywords (\textit{shock/hypotension, infection/sepsis/pneumonia}).
These indicators are transformed through nonlinear mappings to generate both the propensity score and the potential outcome functions:
\[
t_i \sim \text{Bernoulli}(g(\bw_i)), \qquad
y_i = Q_{t_i}(\bw_i) + \varepsilon_i.
\]
This semi-synthetic design yields a known ground truth $\tau_{\text{true}} = \mathbb{E}[Y(1) - Y(0)]$. Details on semi-synthetic data construction are provided in Appendix~\ref{sec:synthetic_data}. 
 
\textbf{Baselines.} We compare our method against \rev{five} baseline methods: TARNet, CFRNet~\citep{shalit2017estimating}, DragonNet~\citep{shi2019adapting}, CausalBERT~\citep{pmlr-v124-veitch20a}, \rev{and DIVA~\citep{zhou2023causal}}. 
For all models, we use the officially released source code when available, or reimplement them using the optimal hyperparameter settings reported in their respective papers. 
To ensure comparability, all text-based models use the same encoder backbone, ClinicalBERT~\citep{wang2023optimized, liu2025generalist}. Implementation uses the Adam optimizer with early stopping based on validation loss. 
Data are randomly split into training, validation, and test sets in a 25:6:6 ratio, which are used respectively for model fitting, early stopping, and final evaluation. 
See more details in Appendix~\ref{sec:implementation}.

\begin{table}[!t]
\footnotesize
\setlength{\tabcolsep}{4pt}
\centering
\caption{
Comparison of ATE estimation and positivity diagnostics across methods.
Results are averaged over 200 replications with bootstrap size $B=1000$.
Confidence intervals use a 95\% confidence level, and propensity scores are clipped at 0.03/0.97.
TARNet and CFRNet do not explicitly estimate propensity scores, so only OR estimation is reported.
}
\label{tab:simulation}

\begin{tabular}{llrrrrrr}
\toprule
\multirow{2}{*}{Method} & \multirow{2}{*}{Estimator}
& \multicolumn{4}{c}{ATE Metrics}
& \multicolumn{2}{c}{PS Quality} \\
\cmidrule(lr){3-6}\cmidrule(lr){7-8}

&  & Mean$(|$Bias$|)$ 
& SD$(\widehat{\text{ATE}})$ 
& Avg SE 
& C.I. Coverage 
& ESS Ratio 
& Clip. Frac. \\
\midrule

TARNet
& OR  & 0.145 & 0.178 & 0.022 & 21.0\% & -- & -- \\
\midrule

CFRNet
& OR & 0.143 & 0.173 & 0.022 & 20.5\% & -- & -- \\
\midrule

\multirow{3}{*}{DragonNet}
& OR  & 0.160 & 0.199 & 0.020 & 16.5\% 
& \multirow{3}{*}{0.272} 
& \multirow{3}{*}{28.89\%} \\
& IPW  & 1.188 & 1.044 & 0.617 & 35.5\% & & \\
& AIPW & 0.320 & 0.438 & 0.329 & 85.0\% & & \\
\midrule

\multirow{3}{*}{\rev{DIVA}}
& \rev{OR}   & \rev{0.313} & \rev{0.365} & \rev{0.024} & \rev{8.0\%}
& \multirow{3}{*}{\rev{0.340}}
& \multirow{3}{*}{\rev{1.47\%}} \\
& \rev{IPW}  & \rev{0.936} & \rev{1.157} & \rev{0.453} & \rev{56.5\%} & & \\
& \rev{AIPW} & \rev{0.470} & \rev{0.620} & \rev{0.288} & \rev{72.0\%} & & \\
\midrule

\multirow{3}{*}{CausalBERT}
& OR  & 0.194 & 0.243 & 0.023 & 13.5\%
& \multirow{3}{*}{0.266}
& \multirow{3}{*}{30.96\%} \\
& IPW  & 1.393 & 1.104 & 0.649 & 48.5\% & & \\
& AIPW & 0.351 & 0.475 & 0.349 & 87.5\% & & \\
\midrule

\multirow{3}{*}{CATR}
& OR  & 0.186 & 0.231 & 0.016 & 10.0\%
& \multirow{3}{*}{0.310}
& \multirow{3}{*}{10.74\%} \\
& IPW  & {0.704}
& {0.845}
& {0.494}
& {76.0\%}
& & \\
& AIPW & {0.261}
& {0.317}
& {0.265}
& {90.0\%}
& & \\

\bottomrule
\end{tabular}
\end{table}

\textbf{Evaluation Protocol and Metrics.} Each model produces an estimated propensity score $\widehat g(\bw)$ and conditional outcomes $\widehat Q_t(\bw)$, which are then used to compute the ATE estimates, as defined in Equations~(\ref{eq:or})-(\ref{eq:aipw}). 
Each model is evaluated over $R=200$ independent replications, each producing three ATE estimates based on newly sampled treatment and outcome assignments. 
For each replication, we compute the point estimate $\widehat{\tau}_{r}$ and its bootstrap standard deviation $\widehat{\mathrm{SE}}_{r}$ using $B=1000$ resamples, and construct the 95\% confidence interval $[\widehat{\tau}_{r} \pm 1.96\,\widehat{\mathrm{SE}}_{r}]$. 
Detailed definitions of the evaluation metrics are provided in Appendix~\ref{sec:implementation}. 
For each method, the ESS ratio and clipping fraction are reported once, averaged across all replications, while ATE metrics are reported separately for OR, IPW, and AIPW estimators.
% A lower mean absolute bias and standard deviation, together with higher coverage and ESS ratio, reflect more accurate and stable causal effect estimation with improved treatment overlap.

Note that OR yields unrealistically small standard errors compared with the corresponding empirical standard deviations. 
\rev{A likely reason is that OR forms counterfactuals entirely from the fitted outcome regressions and must extrapolate to units with limited support in the opposite group. 
A standard outcome model is unable to signal the extrapolation and produces narrow intervals with severe undercoverage. 
IPW and AIPW instead incorporate propensity scores, so the treatment imbalance is accounted for.} 

Because OR estimates are overconfident and unreliable in this setting, we focus on IPW and AIPW. As shown in Table~\ref{tab:simulation}, CATR substantially improves estimation stability compared to all baselines. 
By reducing extreme propensity scores and increasing the effective sample size, it mitigates observational-level positivity violations. 
This leads to lower bias and variance for both IPW and AIPW estimators, as well as improved confidence interval coverage, particularly for IPW, which relies entirely on propensity score estimation. 
In contrast, DragonNet and CausalBERT produce unstable inverse weights and wider confidence intervals. 
Overall, these results suggest that HSIC-guided sparse rationalization improves statistical efficiency while preserving essential confounding signals. 
 
The ablation study in Table~\ref{tab:ablation} further demonstrates the effectiveness of CATR. 
For IPW estimation, removing the HSIC term substantially increases bias, variance, and clipping frequency, highlighting the importance of the residual-independence constraint for stable propensity estimation. 
Removing both HSIC and sparsity leads to even larger degradation, with severe IPW instability and poor confidence interval behavior, illustrating the role of CATR in mitigating positivity violations. For AIPW estimation, CATR achieves comparable or higher coverage rates while simultaneously reducing both standard deviation and standard error, indicating more stable inference and improved uncertainty quantification. 
Overall, these results show that CATR consistently improves estimation stability for propensity-weighted estimators.

\begin{table}[!t]
\footnotesize
\setlength{\tabcolsep}{4pt}
\renewcommand{\arraystretch}{1.12}
\centering

\caption{
CATR ablation study.
Results are averaged over 200 replications with bootstrap size $B=1000$.
Confidence intervals use a 95\% confidence level, and propensity scores are clipped at 0.03/0.97.
\rev{Both ablations retain the token-selection module and remove only the penalty terms to separate the effect of the selection architecture itself from that of the residual-HSIC and sparsity criteria.}
}
\label{tab:ablation}

\begin{tabular}{llrrrrrr}
\toprule
\multirow{2}{*}{Method} & \multirow{2}{*}{Estimator}
& \multicolumn{4}{c}{ATE Metrics}
& \multicolumn{2}{c}{PS Quality} \\
\cmidrule(lr){3-6}\cmidrule(lr){7-8}

& & Mean$(|$Bias$|)$
& SD$(\widehat{\text{ATE}})$
& Avg SE
& C.I. Coverage
& ESS Ratio
& Clip. Frac. \\
\midrule

\multirow{3}{*}{Full CATR}
& OR   & 0.186 & 0.231 & 0.016 & 10.0\%
& \multirow{3}{*}{0.310}
& \multirow{3}{*}{10.74\%} \\
& IPW  & 0.704 & 0.845 & 0.494 & 76.0\% & & \\
& AIPW & 0.261 & 0.317 & 0.265 & 90.0\% & & \\
\midrule

\multirow{3}{*}{$-$HSIC}
& OR   & 0.150 & 0.186 & 0.015 & 10.5\%
& \multirow{3}{*}{0.290}
& \multirow{3}{*}{17.36\%} \\
& IPW  & 0.978 & 0.947 & 0.570 & 67.5\% & & \\
& AIPW & 0.278 & 0.335 & 0.304 & 90.0\% & & \\
\midrule

\multirow{3}{*}{\makecell[l]{$-$HSIC \\ $-$Sparsity}}
& OR   & 0.161 & 0.199 & 0.020 & 17.0\%
& \multirow{3}{*}{0.270}
& \multirow{3}{*}{29.55\%} \\
& IPW  & 1.241 & 1.083 & 0.623 & 51.0\% & & \\
& AIPW & 0.321 & 0.438 & 0.333 & 89.5\% & & \\

\bottomrule
\end{tabular}
\end{table}

Appendix Figure~\ref{fig:hsic_importance} compares the top-20 important words selected with and without HSIC regularization. 
Without HSIC regularization, most selected tokens appear only once and are largely unrelated to the pre-specified confounding words, suggesting that the model captures spurious correlations rather than meaningful causal signals. 
In contrast, incorporating HSIC successfully recovers all manually defined confounding words (\textit{sepsis, infection, pneumonia, shock, hypotension}) together with related variants (e.g., \textit{shocked}, \textit{pneumonic}). 
By penalizing residual dependence between $T$ and $Y$, HSIC helps the model avoid coincidental co-occurrences and improves the interpretability of the learned representations.
 
\revv{Additional experiments examine alternative confounding structures and heterogeneous treatment effects.
Appendix~\ref{app:dgp} shows that CATR can reduce IPW/AIPW bias under sparse confounding, but exhibits higher bias than the no-selection model under whole-document confounding, highlighting CATR's suitability for settings where adjustment-relevant information is concentrated in a relatively sparse portion of the text.
Appendix~\ref{app:pehe} further reports performance under heterogeneous effects, showing that CATR is able to maintain its CATE accuracy as overlap deteriorates.}

\subsection{Real Data Analysis}
\label{sec:real}

To complement the semi-synthetic experiments, we analyze the same cohort from the MIMIC-III database as in the simulation, with naturally occurring treatment and outcomes. 
The treatment is defined as intravenous (IV) fluid therapy and the outcome is defined as ICU readmission. 
For each patient, we extract both physician notes and structured covariates (40 variables including demographics, comorbidities, vitals, and labs, provided in Appendix~\ref{sec:structured_vars}). 
This setting also evaluates our multimodal extension (Section~\ref{subsec:extension}), which integrates structured variables and unstructured text to model confounding. 
% The analysis allows us to assess whether our framework remains effective in practical applications with real treatment assignments and outcomes. 

\begin{table}[!t]
\footnotesize
\setlength{\tabcolsep}{4pt}
\renewcommand{\arraystretch}{1.12}
\centering

\caption{
Comparison of ATE estimation and positivity diagnostics on real MIMIC-III datasets. Means and standard errors are calculated across 20 replications. 
The structured-only arm uses logistic regression and is deterministic, so no standard error is reported. 
}
\label{tab:real_data}

\begin{tabular}{lccccc}
\toprule
& \multicolumn{3}{c}{ATE (SE)}
& \multicolumn{2}{c}{PS Quality} \\
\cmidrule(lr){2-4}
\cmidrule(lr){5-6}

& OR
& IPW
& AIPW
& ESS Ratio
& Clip. Frac. \\
\midrule

Structured-only
& $-0.069$
& $-0.000$
& $-0.043$
& 0.211
& 8.18\% \\

\rev{Text-only}
& \rev{$0.039$ $(0.011)$}
& \rev{$0.899$ $(0.190)$}
& \rev{$-0.198$ $(0.087)$}
& \rev{$0.201$}
& \rev{$25.64\%$} \\

Multi-modal (w/o CATR)
& $0.060$ $(0.009)$
& $0.744$ $(0.169)$
& $-0.230$ $(0.085)$
& $0.216$ 
& $24.02\%$ \\

Multi-modal (w/ CATR)
& ${-0.048}$ $(0.013)$
& ${0.019}$ $(0.021)$
& ${-0.028}$ $(0.006)$
& ${0.827}$ 
& ${0.00\%}$ \\

\bottomrule
\end{tabular}
\end{table}

We compare \rev{four} adjustment strategies: structured-only, \rev{text-only,} multimodal without CATR, and multimodal with CATR. The structured-only setting uses logistic regression for propensity and outcome modeling, while the \rev{text-only and} multimodal settings follow the architecture in Section~\ref{subsec:extension}. 

As shown in Table~\ref{tab:real_data}, adding unfiltered text raises the clipping fraction, illustrating the observational-level positivity violations induced by raw text embeddings. 
\rev{The text-only and multi-modal arms without CATR behave almost identically on every diagnostic. In our setting, where the clinically salient structured indicators are typically also described in the physician notes, the benefit of CATR mainly comes from using text more efficiently.} 
The multimodal model with CATR yields the most stable ATE estimates across OR, IPW, and AIPW, together with higher ESS and lower clipping fractions. 
In addition, the estimated effects become more consistent across estimators. 
\rev{Because the true effect is unknown here, we read these results as evidence of improved overlap and estimator stability rather than of greater accuracy.}
 
Beyond quantitative results, Appendix Figure~\ref{fig:mimic3} shows that CATR selects clinically meaningful tokens, such as terms indicating infections, organ dysfunction, or other severity markers that influence both treatment decisions and outcomes. These include keywords like ``sepsis'', ``infection'', and ``hypotension'', as well as descriptors of patient condition (e.g., ``labored breathing'', ``unsure''). 

% In summary, CATR improves empirical overlap, stabilizes propensity weighting, and enhances interpretability without relying on ad hoc trimming. When combined with structured covariates, these benefits are further amplified, producing narrower confidence intervals and more robust ATE estimates.

%% file: Sections/6-conclusion.tex
\section{Conclusion}\label{sec:con}
We propose a framework for causal effect estimation when text is used to adjust for confounding.
Our approach explicitly targets \emph{observational-level positivity violations}, which occur when high-dimensional textual confounders induce near-deterministic treatment assignment despite positivity holding at the latent level. 
To address this, we incorporate selective rationalization that \rev{learns a compact adjustment representation supporting nuisance prediction with low residual dependence}, yielding representations with improved overlap. 
Empirically, this design reduces propensity score extremity and stabilizes IPW and AIPW estimation, \revv{particularly when confounding-relevant information is concentrated in a relatively sparse subset of the text}. The effectiveness of our method highlights the practical value of positivity-aware rationalization for causal inference from natural language.
% Despite these contributions, our method has limitations.  It assumes that sufficient confounding information can be distilled into a sparse set of tokens,  which may not hold in domains with diffuse or high-entropy signals. 
% Moreover, like all observational approaches, it remains vulnerable if the text variable does not cover all confounding information. 

Future work could strengthen causal interpretability by integrating counterfactual reasoning to \revv{better assess the} necessity and sufficiency of selected tokens. 
In addition, the framework can be naturally combined with post-hoc adjustment methods, such as temperature scaling, beta or histogram calibration, and covariate balancing techniques, to further stabilize propensity estimates and reduce residual bias. 
%More broadly, we hope our framework encourages further investigation of how high-dimensional text proxies interact with core causal assumptions such as positivity and unconfoundedness, providing opportunities to refine text-based causal inference in applied domains, including healthcare and policy.

\section*{Acknowledgements}
The authors thank the Action Editor and anonymous reviewers for their constructive and insightful feedback.
This work was supported by the National Science Foundation under grant DMS-CDS\&E-MSS No. 2401271.

%% file: Sections/8-appendix.tex
\newpage
\section{Proofs and Theoretical Details}
\label{sec:sproof}

\subsection{Assumptions}

For convenience, we first restate the assumptions following their order in the main paper.

\begin{restatedassumption}{assumption:unconfoundedness}{Confounding Sufficiency}
The true confounding signals $\bW^c$ (contained within the
observed text $\bW$) are sufficient for adjustment:
\[
\{Y(0),Y(1)\}\perp T \mid \bW^c .
\]
\end{restatedassumption}

\begin{restatedassumption}{assumption:positivity}{Positivity on True Signals}
    Each unit has a nonzero probability of receiving both treatment and control conditional on the true confounding signals $\bW^c$:
    \[0 < p_1 \leq \Pr(T=1 \mid \bW^c)\leq p_2 < 1, \forall  \bW^c.\]
\end{restatedassumption}

\begin{restatedassumption}{assumption:consistency}{Consistency}
    The observed outcome corresponds to the potential outcome under the received treatment: $Y=Y(T)$.
\end{restatedassumption}

\begin{restatedassumption}{assumption:noise}{Noise Structure}
There exist measurable functions $g^\star:\mathcal{W}^c \to [0,1]$ and $Q_T^\star:\{0,1\}\times\mathcal{W}^c \to \mathbb{R}$ such that
\rev{
\(
T \mid \bW^c \sim \mathrm{Bernoulli}\bigl(g^\star(\bW^c)\bigr),\ 
Y = Q_T^\star(\bW^c) + \varepsilon_Y,
\)
and the outcome noise $\varepsilon_Y$ satisfies
(i) $\varepsilon_Y \perp (T,\bW)$ and
(ii) $\E[\varepsilon_Y] = 0$, $\E[\varepsilon_Y^2] < \infty$.}
\end{restatedassumption}
\begin{restatedassumption}{assumption:data}{Regularity of data and representation}
The treatment $T\in\{0,1\}$ and outcome $Y\in[-M,M]$ for some $M>0$. 
The observed text $\bW$ is encoded by a pretrained model 
into a bounded continuous feature vector $\mathbf{x}=\Lambda(\bW)\in[-1,1]^d$, 
whose distribution admits a bounded density on $[-1,1]^d$, 
where $d$ is the encoder output dimension.
\end{restatedassumption}

\begin{restatedassumption}{assumption:smoothness}{Smoothness}
For each target function $f\in\{g,Q_0,Q_1\}$,
assume that $f$ lies in a Sobolev ball
$\mathcal{W}^{\beta,\infty}([-1,1]^d)$ with smoothness $\beta>0$, i.e.,
\[
f(\mathbf{x})\in\mathcal{W}^{\beta,\infty}([-1,1]^d)
:=\Bigl\{
f:\max_{|\alpha|\le\beta}\mathrm{ess\,sup}_{\mathbf{x}\in[-1,1]^d}
|D^{\alpha}f(\mathbf{x})|\le 1
\Bigr\},
\]
where $\alpha=(\alpha_1,\ldots,\alpha_d)$, $|\alpha|=\sum_j \alpha_j$,
and $D^{\alpha}f$ denotes the weak derivative.
\end{restatedassumption}

\subsection{Proof of Proposition~\ref{prop:hsic}}
We restate the result for convenience.  
For any candidate subset $\widetilde{\bW} \subseteq \bW$, define
\[
g(\widetilde{\bW}) = \E[T \mid \widetilde{\bW}],
\qquad
Q_t(\widetilde{\bW}) = \E[Y \mid T = t, \widetilde{\bW}],
\]
and residuals
\[
r_T = T - g(\widetilde{\bW}),
\qquad
r_Y = Y - Q_T(\widetilde{\bW}).
\]
Let
\(
D(\widetilde{\bW}) = \mathrm{HSIC}(r_T, r_Y)
\)
denote the population-level HSIC between the treatment and outcome residuals, computed with characteristic and bounded kernels, under finite second moments.  
Under Assumptions~\ref{assumption:unconfoundedness},~\ref{assumption:consistency}, and~\ref{assumption:noise}, we show that
\[
\revv{\left.
\begin{aligned}
(Y(0), Y(1)) &\perp T \mid \widetilde{\bW},\\
Q_t^\star(\bW^c) &= \widetilde Q_t(\widetilde{\bW}),\quad t=0,1
\end{aligned}
\right\}
\quad\Longrightarrow\quad
D(\widetilde{\bW}) = 0.}
\]

\paragraph*{Step 1. Residuals recover exogenous noise under sufficiency}

By the consistency assumption, $Y = Y(T)$.  
\revon
Assumption~\ref{assumption:unconfoundedness} states that
$(Y(0), Y(1)) \perp T \mid \bW^c$, and we suppose that $\widetilde{\bW}$ is sufficient in the
sense of Proposition~\ref{prop:hsic}, that is,
\[
(Y(0), Y(1)) \perp T \mid \widetilde{\bW},
\qquad
Q^\star_t(\bW^c) = \widetilde Q_t(\widetilde{\bW}),\quad t\in\{0,1\}.
\]
The argument below uses the second condition, which the submitted proof invoked implicitly:
conditional unconfoundedness constrains the treatment assignment given $\widetilde{\bW}$, but
does not by itself require $\widetilde{\bW}$ to determine the outcome mechanism.
\normalcolor

\revon
The treatment residual requires no condition on $g^\star$. By definition
$g(\widetilde{\bW}) = \E[T\mid\widetilde{\bW}]$ is a function of $\widetilde{\bW}$, and
$\widetilde{\bW}$ is a function of $\bW$, so
\[
r_T = T - g(\widetilde{\bW})
\]
is a function of $(T,\bW)$. In particular no additive representation of the binary treatment
is needed.
\normalcolor

Under the additive-noise outcome model in Assumption~\ref{assumption:noise},
\[
Y = Q_T^\star(\bW^c) + \varepsilon_Y,
\qquad
\varepsilon_Y \perp (T, \bW),
\qquad
\E[\varepsilon_Y] = 0,\;\E[\varepsilon_Y^2] < \infty.
\]
\rev{Retention of the outcome mechanism gives $Q^\star_t(\bW^c)=\widetilde Q_t(\widetilde{\bW})$, and $\E[\varepsilon_Y\mid T=t,\widetilde{\bW}]=0$ since $\varepsilon_Y\perp(T,\bW)$; hence}
\[
Q_t(\widetilde{\bW}) = \E[Y \mid T = t, \widetilde{\bW}] = \E[Q_T^\star(\bW^c) + \varepsilon_Y \mid T = t, \widetilde{\bW}]
                 = Q_t^\star(\bW^c),
\]
and hence
\[
r_Y = Y - Q_T(\widetilde{\bW})
    = \bigl(Q_T^\star(\bW^c) + \varepsilon_Y\bigr) - Q_T^\star( \bW^c)
    = \varepsilon_Y.
\]

\rev{Thus, under sufficiency of $\widetilde{\bW}$, the outcome residual coincides with the
exogenous noise, $r_Y = \varepsilon_Y$, while the treatment residual is a function of $(T,\bW)$.}

\paragraph*{Step 2. Independence of residuals}

\revon
By Assumption~\ref{assumption:noise}, $\varepsilon_Y \perp (T,\bW)$. Since $r_T$ is a function of $(T,\bW)$ and independence is preserved under such transformations,
\(
\varepsilon_Y \perp r_T .
\)
Combining this with $r_Y = \varepsilon_Y$ from Step 1 gives
\(
r_T \perp r_Y .
\)
\normalcolor

\paragraph*{Step 3. HSIC characterization of independence}

For bounded characteristic kernels $k_T(\cdot,\cdot)$ and $k_Y(\cdot,\cdot)$,
the Hilbert--Schmidt independence criterion satisfies
\[
\mathrm{HSIC}(U, V) = 0
\quad\Longleftrightarrow\quad
U \perp V,
\]
for any random variables $U, V$ with finite second moments
\citep{gretton2005measuring,gretton2007kernel}.  
Applying this to $U = r_T$ and $V = r_Y$ yields
\[
r_T \perp r_Y
\quad\Longrightarrow\quad
\mathrm{HSIC}(r_T, r_Y) = 0.
\]

\paragraph*{Step 4. Conclusion}

\revv{Combining Steps~1--3, we conclude that if $\widetilde{\bW}$ satisfies
both sufficiency conditions in Proposition~\ref{prop:hsic}, then the treatment
and outcome residuals are independent and their HSIC equals zero:}
\[
\revv{\left.
\begin{aligned}
(Y(0), Y(1)) &\perp T \mid \widetilde{\bW},\\
Q_t^\star(\bW^c) &= \widetilde Q_t(\widetilde{\bW}),\quad t=0,1
\end{aligned}
\right\}
\quad\Longrightarrow\quad
D(\widetilde{\bW}) = \mathrm{HSIC}(r_T, r_Y) = 0.}
\]
The contrapositive follows immediately under the structural additive-noise model:
if $D(\widetilde{\bW}) > 0$, then $r_T$ and $r_Y$ are dependent, and $\widetilde{\bW}$
\revv{cannot satisfy both sufficiency conditions in Proposition~\ref{prop:hsic}.} \hfill $\square$

\subsection{Proof of Theorems~\ref{thm:penalized},~\ref{thm:consistency}, and~\ref{thm:ipw-aipw}}

We restate the result here for completeness. 
Under Assumptions~\ref{assumption:data}--\ref{assumption:smoothness},
consider the joint penalized objective
\begin{equation*}
\min_{\theta,\phi}\;
\frac{1}{n}\sum_{i=1}^n 
\Big[
\mathcal{L}_{\text{sup}, i} 
+ \mu\,\mathcal{R}_{\text{sparse}}(s_i)
+ \gamma\,\mathrm{HSIC}(r_{T,i},r_{Y,i})
\Big],
\end{equation*}
where $\mathcal{L}_{\text{sup},i}$ is the supervised prediction loss,
$\mathcal{R}_{\text{sparse}}$ is the sparsity regularizer,
and $\mathrm{HSIC}(r_{T,i},r_{Y,i})$ measures residual dependence.
Let $(\hat\theta,\hat\phi)$ be the minimizer, and denote
$\hat f$ as either $\hat g$ or $\hat Q_t$.
\rev{For penalty weights $\mu_n,\gamma_n>0$ that may depend on $n$, for an MLP-ReLU network with width 
$H\asymp n^{\frac{d}{2(\beta+d)}}\log^2 n$
and depth $D\asymp\log n$, there exists $C>0$ such that, with probability at least 
$1-\exp(-n^{\frac{d}{\beta+d}}\log^8 n)$,
\[
\|\hat f - f\|_{L_2(P_X)}^2
\le
C\bigl(a_n + b_n\bigr),
\qquad
a_n = n^{-\frac{\beta}{\beta+d}}\log^8 n + \frac{\log\log n}{n},
\quad
b_n = \mu_n C_{\mathrm{sp}} + \gamma_n C_{\mathrm{HSIC}} .
\]}%
Following Theorem 1 in~\citet{Farrell_2021}, let $\mathcal{F}$ denote the class of MLP-ReLU networks with width $H$ and depth $D$. Under Assumptions~\ref{assumption:data}--\ref{assumption:smoothness},
for an MLP-ReLU network with width 
$H\asymp n^{\frac{d}{2(\beta+d)}}\log^2 n$
and depth $D\asymp\log n$, there exists $C>0$ such that, with probability at least 
$1-\exp(-n^{\frac{d}{\beta+d}}\log^8 n)$,
\[
\|\hat f - f\|_{L_2(P_X)}^2\le C\Bigl(
n^{-\frac{\beta}{\beta+d}}\log^8 n
+\frac{\log\log n}{n}
\Bigr),
\]
which gives the error bound for unrestricted function class.

\revon
The penalty $\mathcal{R}_{\text{sparse}}$ is bounded (the entropy penalty lies in $[0,\log 2]$) and
$\mathrm{HSIC}$ is bounded by the product of kernel norms; write $C_{\mathrm{sp}}$ and
$C_{\mathrm{HSIC}}$ for these bounds. The penalized objective then differs from the unpenalized
one by at most
\[
\Big|
\frac{1}{n}\sum_{i=1}^n
\big[
\mu_n\,\mathcal{R}_{\text{sparse}}(s_i)
+\gamma_n\,\mathrm{HSIC}(r_{T,i},r_{Y,i})
\big]
\Big|
\;\le\;
\mu_n C_{\mathrm{sp}} + \gamma_n C_{\mathrm{HSIC}}
\;=\; b_n .
\]
Boundedness alone therefore gives only $b_n = O(1)$ when the weights are held fixed, which is not
negligible: the penalized minimizer need not converge to $f_*$. \revv{Let $\widehat R_n$ denote the empirical supervised risk and
$P_n$ the total regularization penalty, with $0\le P_n(f)\le b_n$.
Let $f_n$ be the network approximation used in Appendix A.1
of \citet{Farrell_2021}. By optimality of $\widehat f$,
\[
\widehat R_n(\widehat f)+P_n(\widehat f)
\le \widehat R_n(f_n)+P_n(f_n),
\]
and hence
\[
\widehat R_n(\widehat f)\le \widehat R_n(f_n)+b_n.
\]
Under the same loss and network-class conditions, their localization
argument yields}
\[
\revv{\|\widehat f-f_*\|_{L_2(P_X)}^2=O_P(a_n+b_n).}
\]
\revv{A sufficient condition for consistency is $\mu_n,\gamma_n\to0$, so that $b_n=o(1)$,} and the unpenalized rate
$a_n$ is preserved whenever $b_n = O(a_n)$.
\normalcolor

\revv{The empirical residual bound in
Theorem~\ref{thm:consistency}(iii) follows by adapting
the localization argument in Lemma 10 of \citet{Farrell_2021}.} Given the nonasymptotic $L_2$ bounds of the MLP approximation error,
the convergence rate of nuisance estimators, IPW and AIPW estimators of ATE, and the asymptotic normality of the AIPW estimator directly follow from classical semiparametric theory, as established in Theorem 3 of \citep{Farrell_2021}.
Therefore, we omit the proof for brevity.

\section{Semi-Synthetic Data Construction}
\label{sec:synthetic_data}
We synthesize treatments and outcomes from text-derived binary confounders following a standard potential-outcomes framework. 

\paragraph*{Confounders from text}
For each clinical note, we extract two binary indicators using simple keyword matching:
\[
C_i = (\texttt{has\_shock}_i, \texttt{has\_infection}_i) \in \{0,1\}^2,
\]
where $\texttt{has\_shock}_i = 1$ if shock-related terms appear in the text, and $\texttt{has\_infection}_i = 1$ if infection- or sepsis-related terms appear.

\paragraph*{Propensity scores and treatment assignment}
To produce the propensity score $g_i$, each combination of $(\texttt{has\_shock}, \texttt{has\_infection})$ corresponds to a pre-specified cell-wise propensity. Treatments are independently drawn as
\(T_i \sim \text{Bernoulli}(g_i).\)
This design introduces confounding through $C \rightarrow T$, while maintaining overlap by bounding $g_i$ away from 0 and 1.

\paragraph*{Outcome model and potential outcomes}
Control potential outcomes are generated via a linear model with additive Gaussian noise:
\begin{align*}
    & Y_i(0) = \beta_0 + \beta_1\cdot \texttt{has\_shock}_i + \beta_2\cdot \texttt{has\_infection}_i + \varepsilon_i,
\quad \varepsilon_i \sim \mathcal{N}(0, \sigma^2),\\
    & {Y}_i(1) = Y_i(0) + \tau_{\text{True}}.
\end{align*}

Our simulation is carefully designed to reflect key challenges in real-world text-based causal inference. The text inputs are actual MIMIC-III clinical notes, preserving the semantic and distributional complexity of real clinical language. We introduce nonlinear relationships between confounders, treatment, and outcome. Together, these elements create a testbed that is both analytically tractable and empirically demanding, making it well-suited for evaluating causal inference methods under realistic yet controlled conditions.

\section{Implementation Details}
\label{sec:implementation}
\paragraph{Selectors and sparsity penalty.} We implement $\bs\in[0,1]^d$ with a temperature-controlled concrete/Gumbel–Softmax gate. For selective rationalization, we use a temperature parameter $\eta = 1.0$ to control the sharpness of the gating distribution. A lower temperature leads to sparser token selection, promoting interpretability and improving stability in high-dimensional settings. The entropy penalty is
\[
\mathcal R_{\mathrm{ent}}(\bs)=\frac{1}{d}\sum_{j=1}^d\big\{-s_j\log s_j-(1-s_j)\log(1-s_j)\big\}.
\]

\paragraph{HSIC penalty.} Given mini-batch residuals $\{e_T^i,e_Y^i\}_{i=1}^B$ computed from current nuisance networks, we compute the empirical HSIC statistic with Linear/Gaussian kernels of bandwidth fixed by the median heuristic per batch. 

\paragraph{Network sizes.} The CATR model uses a pretrained transformer encoder, ClinicalBERT~\citep{wang2023optimized, liu2025generalist}, with hidden size 768, followed by three task-specific heads:

$\bullet$ \textbf{Propensity head:} one hidden layer of dimension 256, ReLU activation, outputting a single logit.\\
$\bullet$ \textbf{Outcome heads ($Q_0, Q_1$):} two hidden layers of width 256 with ReLU activation, each outputting one scalar.\\
$\bullet$ \textbf{Token-importance scorer:} a linear layer.

The total loss is a weighted sum of (i) binary cross-entropy loss for propensity, (ii) binary cross-entropy loss for outcomes, (iii) a sparsity regularizer (L1 or entropy) on token weights, and (iv) an HSIC penalty on residuals:
\[
\mathcal{L} = {\revon \alpha\,}\mathcal{L}_{\text{prop}} + \beta \mathcal{L}_{\text{out}} + \mu \mathcal{L}_{\text{sp}} + \gamma\mathcal{L}_{\mathrm{HSIC}}.
\]
Here \rev{$\alpha$ and} $\beta$ weight the supervised losses, $\mu$ controls sparsity, and $\gamma$ controls the HSIC penalty, consistent with the notation in Equation~\ref{eq:loss}.

\paragraph{Hyperparameter.} In practice, we balance the four loss components (treatment supervision, outcome supervision, sparsity regularization, and HSIC residual decorrelation) so that no single term dominates the objective.
\rev{The weights $(\alpha,\beta,\mu,\gamma)$ are set to $(5,\,1,\,0.1,\,10)$ throughout all experiments. We report them on the scale used in our released code, so that the sensitivity grid of Appendix~\ref{app:sensitivity} is directly comparable; dividing through by $\alpha$ gives the equivalent normalised weights $(1,\,0.2,\,0.02,\,2)$.}

\subsection*{Optimization} We train with Adam using separate learning rates for the encoder and heads, 5e-5 and 2e-5, respectively, and batch size 32. Mixed-precision training (AMP) is used on GPU. 
Early stopping is based on validation loss with patience 3. 

\subsection*{Evaluation Metrics} Across replications, we report:

\begin{itemize}[itemsep=3pt,topsep=0pt,parsep=0pt,partopsep=0pt]
    \item \(\text{Mean}(|\text{Bias}|)\): $\frac{1}{R}\sum_{r=1}^R |\widehat\tau_r - \tau_{\text{true}}|$;
    \item SD\((\widehat{\text{ATE}})\): empirical standard deviation of $\widehat\tau_r$ over replications;
    \item Avg SE: mean of the bootstrap standard deviations, $\frac{1}{R}\sum_{r=1}^R \widehat{\mathrm{SE}}_r$;
    \item C.I. Coverage (95\%): proportion of confidence intervals that cover the true ATE, $\frac{1}{R}\sum_{r=1}^R \mathbb{I}\{\tau_{\text{true}}\in[\widehat\tau_r\pm1.96\,\widehat{\mathrm{SE}}_r]\}$;
    \item ESS Ratio: ratio of the estimated effective sample size to its oracle counterpart, 
    $\mathrm{ESS}_{\text{ratio}} = \mathrm{ESS}(\widehat w)/\mathrm{ESS}(w^\ast)$,
    where $\mathrm{ESS}(w) = (\sum_i w_i)^2 / \sum_i w_i^2$ and $w^\ast$ denotes the true inverse-propensity weights.
    Values closer to 1 indicate better recovery of the effective sample size and stronger empirical overlap;
    \item Clipping Frac.: fraction of units with estimated propensity scores 
    $\widehat g_i\notin[0.03,0.97]$, indicating observational-level positivity violations.
\end{itemize}

\subsection*{Environment and computation resources} PyTorch~2.7.1; CUDA 12.4; Python 3.11. All experiments are conducted on a server with 4 NVIDIA RTX A6000 GPUs.

\subsection*{Open access to code} All code required to reproduce our experiments is provided in the supplementary materials.

\section{Supplementary Experimental Results}

\subsection{\texorpdfstring{\rev{Data-Driven Confounding Designs}}{Data-Driven Confounding Designs}}
\label{app:dgp}

\revon
In the construction of Section~\ref{sec:experiments} the confounding vocabulary is specified by us, which builds the answer into the design. We therefore add three designs in which the confounding signal is derived from an external clinical variable rather than chosen by hand. We use the SOFA (Sequential Organ Failure Assessment) score, a patient-level measure of organ dysfunction, and learn a direction $v$ in the embedding space, using training rows only, such that a note's projection onto $v$ predicts its SOFA score. The scalar confounder $z_i$ is then built in three ways.

\begin{itemize}[itemsep=1pt,topsep=2pt]
\item \emph{Learned sparse lexical.} A Lasso fit predicts the severity projection from binary word-presence indicators; the $43$ words it retains define $z_i$ as their weighted sum in document $i$. The confounding information is concentrated in a sparse vocabulary selected from the data.
\item \emph{Within-region.} Each contextualised token embedding is projected onto $v$ and each document is scanned with $16$-token windows; $z_i$ is the mean projection within the highest-scoring window, on average about $14\%$ of the note. Confounding is diffuse within a short contiguous region rather than tied to particular words.
\item \emph{Whole-document.} $z_i$ is the projection of the full-document embedding onto $v$, so the confounding information is spread over the entire note and corresponds to no token subset. This design lies outside the sparse-confounding premise and is included to locate the boundary of the method.
\end{itemize}

The same $z_i$ then generates treatment and potential outcomes, and the treatment and outcome mechanisms are held fixed across the three designs, so only the locus of confounding varies.

\begin{table}[!t]
\centering
\revon
\footnotesize
\caption{\rev{ATE estimation and overlap under the three data-driven confounding designs, comparing CATR with the no-selection ablation. Entries are means over seeds.}}
\label{tab:dgp}
\begin{tabular}{lcccccc}
\toprule
\multirow{2}{*}{Design}
& \multicolumn{2}{c}{IPW $|$bias$|$ $\downarrow$}
& \multicolumn{2}{c}{AIPW $|$bias$|$ $\downarrow$}
& \multicolumn{2}{c}{ESS $\uparrow$} \\
\cmidrule(lr){2-3}\cmidrule(lr){4-5}\cmidrule(lr){6-7}
& CATR & no selection & CATR & no selection & CATR & no selection \\
\midrule
Learned sparse lexical & \textbf{0.172} & 0.258 & \textbf{0.234} & 0.278 & \textbf{0.826} & 0.773 \\
Within-region          & \textbf{0.276} & 0.394 & \textbf{0.298} & 0.424 & \textbf{0.865} & 0.738 \\
Whole-document         & 0.288 & \textbf{0.134} & 0.388 & \textbf{0.371} & \textbf{0.767} & 0.742 \\
\bottomrule
\end{tabular}
\end{table}

\revon
On the two designs where confounding is carried by a recoverable part of the document, CATR reduces both IPW and AIPW bias while raising the effective sample size. On the whole-document design it does not: the no-selection ablation attains lower IPW and AIPW bias, and only the overlap diagnostic still favours selection. This delimits the intended scope of the method rather than being incidental. When the confounder is a functional of the entire note, there is no sparse subset for the selector to recover, and the selection machinery may add variance. We state this scope condition explicitly rather than claiming uniform gains across all text-confounding structures.
\normalcolor

\subsection{\texorpdfstring{\rev{Heterogeneous Treatment Effects and CATE Estimation}}{Heterogeneous Treatment Effects and CATE Estimation}}
\label{app:pehe}

\revon
The semi-synthetic design of Section~\ref{sec:experiments} fixes $Y_i(1)-Y_i(0)=\tau$, so the true conditional average treatment effect is constant and $\sqrt{\mathrm{PEHE}}$ is uninformative. To evaluate CATE estimation we therefore construct a design with genuinely heterogeneous effects. We use the whole-document confounder $z_i$ of Appendix~\ref{app:dgp}, standardised. Treatment and outcomes are then generated as
\[
T_i \sim \mathrm{Bernoulli}\bigl(0.02 + 0.96\,\sigma(a z_i)\bigr),
\qquad
Y_i(0) = 0.4 z_i + \tanh(z_i) + \varepsilon_i,
\qquad
\tau_i = 2.0 + 0.8 z_i,
\]
with $\varepsilon_i \sim \mathcal N(0,1)$. The true ATE is $2.0$ while the true CATE varies across units. The single parameter $a$ governs how strongly assignment depends on the confounder and therefore controls overlap: raising $a$ widens the propensity range and lowers the ESS ratio, so the same design traces out a range of overlap conditions.

\revon
In Table~\ref{tab:pehe}, we compare against CFRNet and TARNet, the representation-balancing methods most directly aimed at CATE. All methods use the same encoder, the same data splits and the same $20$ seeds. As a reference point we also report the $\sqrt{\mathrm{PEHE}}$ of a constant predictor set to the true ATE, which any method that fails to capture heterogeneity cannot improve upon.

\begin{table}[!t]
\centering
\revon
\footnotesize
\caption{\rev{Held-out $\sqrt{\mathrm{PEHE}}$ under the heterogeneous-effect design, for three overlap regimes indexed by $a$. Entries are means over $20$ seeds with standard errors. The constant-oracle column is the $\sqrt{\mathrm{PEHE}}$ of a predictor that returns the true ATE for every unit.}}
\label{tab:pehe}
\begin{tabular}{cccccc}
\toprule
$a$ & ESS ratio & CATR & CFRNet & TARNet & Constant-oracle \\
\midrule
0.5 & 0.801 & 0.616 {\scriptsize $\pm$ 0.015} & \textbf{0.583} {\scriptsize $\pm$ 0.018} & 0.584 {\scriptsize $\pm$ 0.018} & 0.798 \\
1.2 & 0.524 & \textbf{0.662} {\scriptsize $\pm$ 0.024} & 0.714 {\scriptsize $\pm$ 0.027} & 0.710 {\scriptsize $\pm$ 0.025} & 0.798 \\
2.0 & 0.362 & \textbf{0.633} {\scriptsize $\pm$ 0.017} & 0.890 {\scriptsize $\pm$ 0.030} & 0.882 {\scriptsize $\pm$ 0.030} & 0.798 \\
\bottomrule
\end{tabular}
\end{table}

\revon
The ordering reverses as overlap deteriorates. Under good overlap ($a=0.5$) CFRNet attains the lower mean, and CATR is better at $a=1.2$ and $a=2.0$. CATR's CATE accuracy is essentially flat in $a$ ($0.616$, $0.662$, $0.633$) while the ESS ratio falls from $0.80$ to $0.36$, whereas CFRNet and TARNet degrade monotonically. At $a=2.0$, both CFRNet and TARNet have higher $\sqrt{\mathrm{PEHE}}$
than the constant predictor, indicating that these methods fail to
capture treatment-effect heterogeneity accurately in this setting.

\subsection{\texorpdfstring{\rev{Sensitivity to the Penalty Weights}}{Sensitivity to the Penalty Weights}}
\label{app:sensitivity}

\revon
The weights $\mu$ and $\gamma$ in Equation~\ref{eq:loss} control how strongly the objective favours compact selections and residual independence. We examine the trade-off empirically on the semi-synthetic design of Section~\ref{sec:experiments}. Selection entropy is the normalised entropy of the token weights, so smaller values indicate a more concentrated rationale. AUC is the propensity model's discrimination on the held-out split.

\begin{table}[!t]
\centering
\revon
\footnotesize
\caption{\rev{Sensitivity analysis to the sparsity weight $\mu$ and the HSIC weight $\gamma$. Entries are means over seeds with standard errors; the clipping fraction and ESS ratio are computed as in Section~\ref{sec:implementation}.}}
\label{tab:sensitivity}
\begin{tabular}{rrcccccc}
\toprule
$\mu$ & $\gamma$ & IPW $|$bias$|$ $\downarrow$ & AIPW $|$bias$|$ $\downarrow$ & Clipping $\downarrow$ & ESS $\uparrow$ & Sel.\ entropy & AUC $\uparrow$ \\
\midrule
\multirow{5}{*}{0.00}
 & 0    & 1.207 {\scriptsize $\pm$ 0.196} & 0.290 {\scriptsize $\pm$ 0.062} & 7.2\% & 0.240 & 0.858 & 0.848 \\
 & 1    & 1.250 {\scriptsize $\pm$ 0.205} & 0.302 {\scriptsize $\pm$ 0.062} & 8.4\% & 0.239 & 0.856 & 0.847 \\
 & 10   & 0.968 {\scriptsize $\pm$ 0.177} & 0.290 {\scriptsize $\pm$ 0.048} & 3.8\% & 0.264 & 0.871 & 0.850 \\
 & 100  & 0.585 {\scriptsize $\pm$ 0.106} & 0.206 {\scriptsize $\pm$ 0.053} & 0.5\% & 0.359 & 0.907 & 0.850 \\
 & 500  & 0.892 {\scriptsize $\pm$ 0.188} & 0.834 {\scriptsize $\pm$ 0.138} & 1.0\% & 0.518 & 0.925 & 0.683 \\
\midrule
\multirow{6}{*}{0.10}
 & 0    & 1.087 {\scriptsize $\pm$ 0.199} & 0.208 {\scriptsize $\pm$ 0.044} & 5.1\% & 0.249 & 0.227 & 0.847 \\
 & 1    & 1.197 {\scriptsize $\pm$ 0.207} & 0.201 {\scriptsize $\pm$ 0.045} & 5.8\% & 0.250 & 0.230 & 0.846 \\
 & 10   & 0.788 {\scriptsize $\pm$ 0.134} & 0.279 {\scriptsize $\pm$ 0.034} & 2.4\% & 0.286 & 0.304 & 0.843 \\
 & 100  & {0.487} {\scriptsize $\pm$ 0.092} & 0.250 {\scriptsize $\pm$ 0.034} & {0.4\%} & 0.398 & 0.779 & 0.845 \\
 & 500  & 1.163 {\scriptsize $\pm$ 0.187} & 0.696 {\scriptsize $\pm$ 0.156} & 2.4\% & 0.456 & 0.909 & 0.723 \\
 & 1000 & 0.741 {\scriptsize $\pm$ 0.246} & 1.224 {\scriptsize $\pm$ 0.372} & 7.2\% & 0.572 & 0.903 & 0.605 \\
\midrule
\multirow{4}{*}{1.00}
 & 0    & 1.487 {\scriptsize $\pm$ 0.244} & 0.393 {\scriptsize $\pm$ 0.086} & 9.3\% & 0.239 & 0.010 & 0.848 \\
 & 1    & 1.260 {\scriptsize $\pm$ 0.181} & 0.297 {\scriptsize $\pm$ 0.053} & 7.7\% & 0.246 & 0.010 & 0.850 \\
 & 10   & 1.292 {\scriptsize $\pm$ 0.214} & 0.238 {\scriptsize $\pm$ 0.051} & 6.5\% & 0.252 & 0.013 & 0.848 \\
 & 100  & 0.495 {\scriptsize $\pm$ 0.092} & {0.194} {\scriptsize $\pm$ 0.042} & 1.6\% & {0.408} & 0.049 & 0.845 \\
\bottomrule
\end{tabular}
\end{table}

\revon
Across all three values of $\mu$, increasing $\gamma$ to $100$ substantially improves IPW bias and overlap diagnostics, while leaving the propensity model's discrimination unchanged. Beyond that point the behaviour reverses: AIPW bias deteriorates sharply and AUC falls, even though the ESS ratio increases. An excessively large HSIC weight therefore overemphasises the residual-independence surrogate at the expense of the predictive information that nuisance estimation requires.

\normalcolor

\subsection{\texorpdfstring{\rev{Robustness to the Hard-Selection Rule}}{Robustness to the Hard-Selection Rule}}
\label{app:hardselect}

\revon
At evaluation the selector produces continuous weights, and sometimes we prefer to generate discrete rationales by turning weights into a hard selection. We check whether the causal estimates depend on the rule of hard selection. Over $10$ seeds on the semi-synthetic design, the learned weights are strongly concentrated: Half of the total weight falls on $6.4$ tokens, and the ten largest weights account for $85.5\%$ of the total.

\revon
We compare four families of post-hoc hard selection against the soft-selected representation used for the main results in Table~\ref{tab:hardselect}. The rules $w_i>1/L$ and $w_i>2/L$ keep tokens whose weight exceeds one or two times the uniform baseline $1/L$ for a document of length $L$; Nucleus $90\%$ keeps the smallest set of highest-weight tokens whose cumulative weight reaches $0.9$; Top $20$ and Top $50$ keep the tokens with the largest weights. For each rule the retained weights are renormalised to sum to one before the pooled representation and the downstream estimates are recomputed.

\begin{table}[!t]
\centering
\revon
\footnotesize
\caption{\rev{Downstream estimates under post-hoc hard-selection rules, averaged over $10$ seeds. ``Tokens kept'' is the mean number of retained tokens per document. The soft-selection row retains all tokens with their learned weights and is the setting used for the main results.}}
\label{tab:hardselect}
\begin{tabular}{lccccccc}
\toprule
Rule & Tokens kept & OR $|$bias$|$ $\downarrow$ & IPW $|$bias$|$ $\downarrow$ & AIPW $|$bias$|$ $\downarrow$ & ESS $\uparrow$ & Clipping $\downarrow$ & AUC $\uparrow$ \\
\midrule
Soft (all)     & 262.8 & \textbf{0.174} & 0.782 & 0.249 & 0.290 & 4.1\% & 0.841 \\
$w_i>1/L$      & 16.1  & 0.194 & 0.767 & 0.247 & 0.303 & 4.0\% & 0.842 \\
$w_i>2/L$      & 9.0   & 0.203 & 0.778 & \textbf{0.245} & \textbf{0.309} & \textbf{4.0\%} & \textbf{0.845} \\
Nucleus $90\%$ & 31.9  & 0.185 & 0.807 & 0.255 & 0.291 & 4.4\% & 0.840 \\
Top $20$       & 20.0  & 0.189 & \textbf{0.761} & 0.251 & 0.305 & 4.0\% & 0.843 \\
Top $50$       & 50.0  & 0.183 & 0.762 & 0.248 & 0.300 & 4.1\% & 0.843 \\
\bottomrule
\end{tabular}
\end{table}

\revon
The estimates are stable across all rules. Because the weight mass is concentrated on a few tokens, any reasonable rule retains essentially the same high-weight information. We therefore keep soft selection for the reported causal estimates.
\normalcolor

\subsection{Selected Tokens}

\begin{figure}[t]
    \centering
    \includegraphics[width=0.85\linewidth]{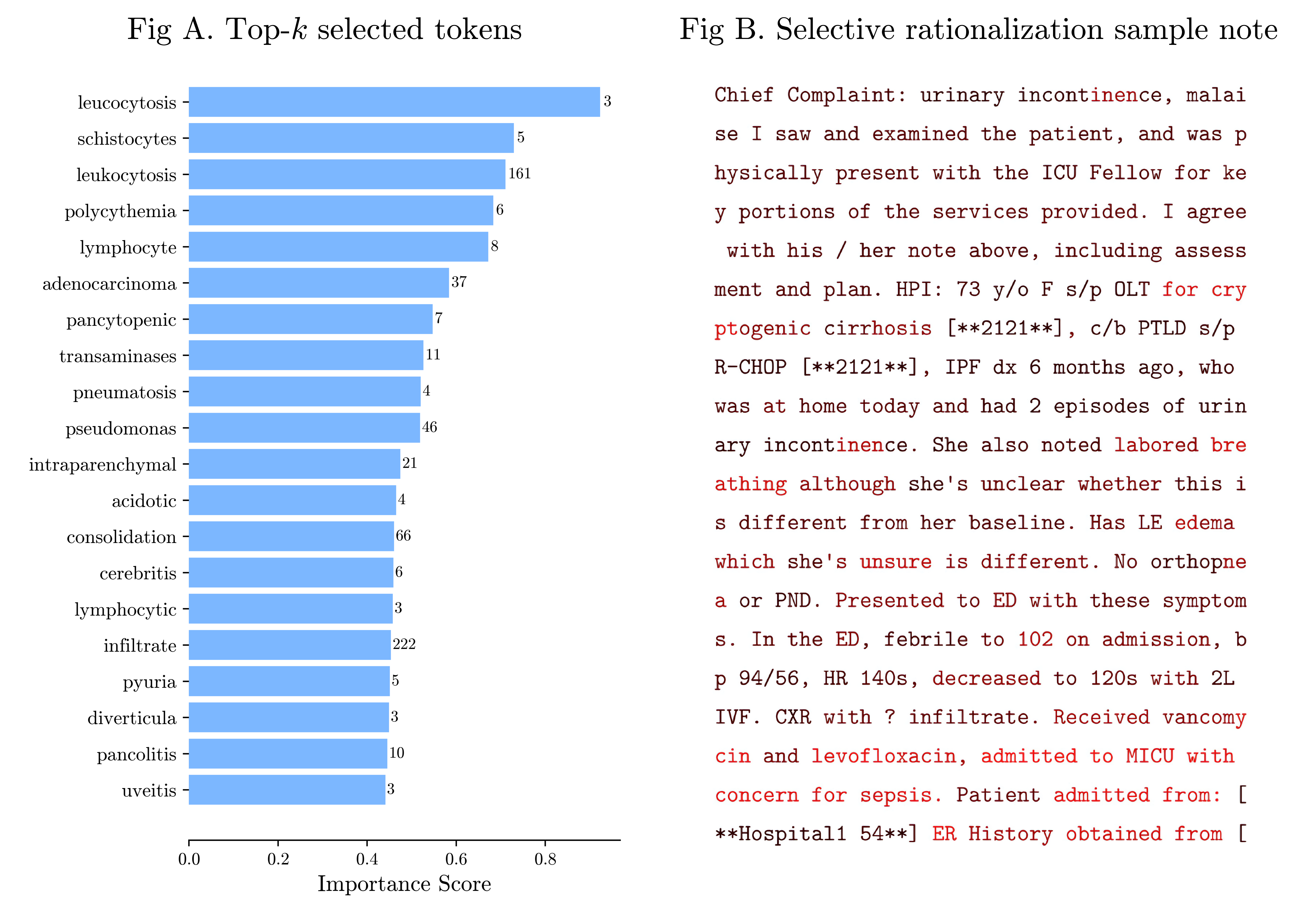}
    \caption{Fig.~A shows top-$k$ selected tokens that are medically informative. 
        Fig.~B visualizes the selective rationalization output with color-encoded selection importance scores.}
    \label{fig:mimic3}
\end{figure}

\begin{figure}[t]
    \centering
    \includegraphics[width=0.85\linewidth]{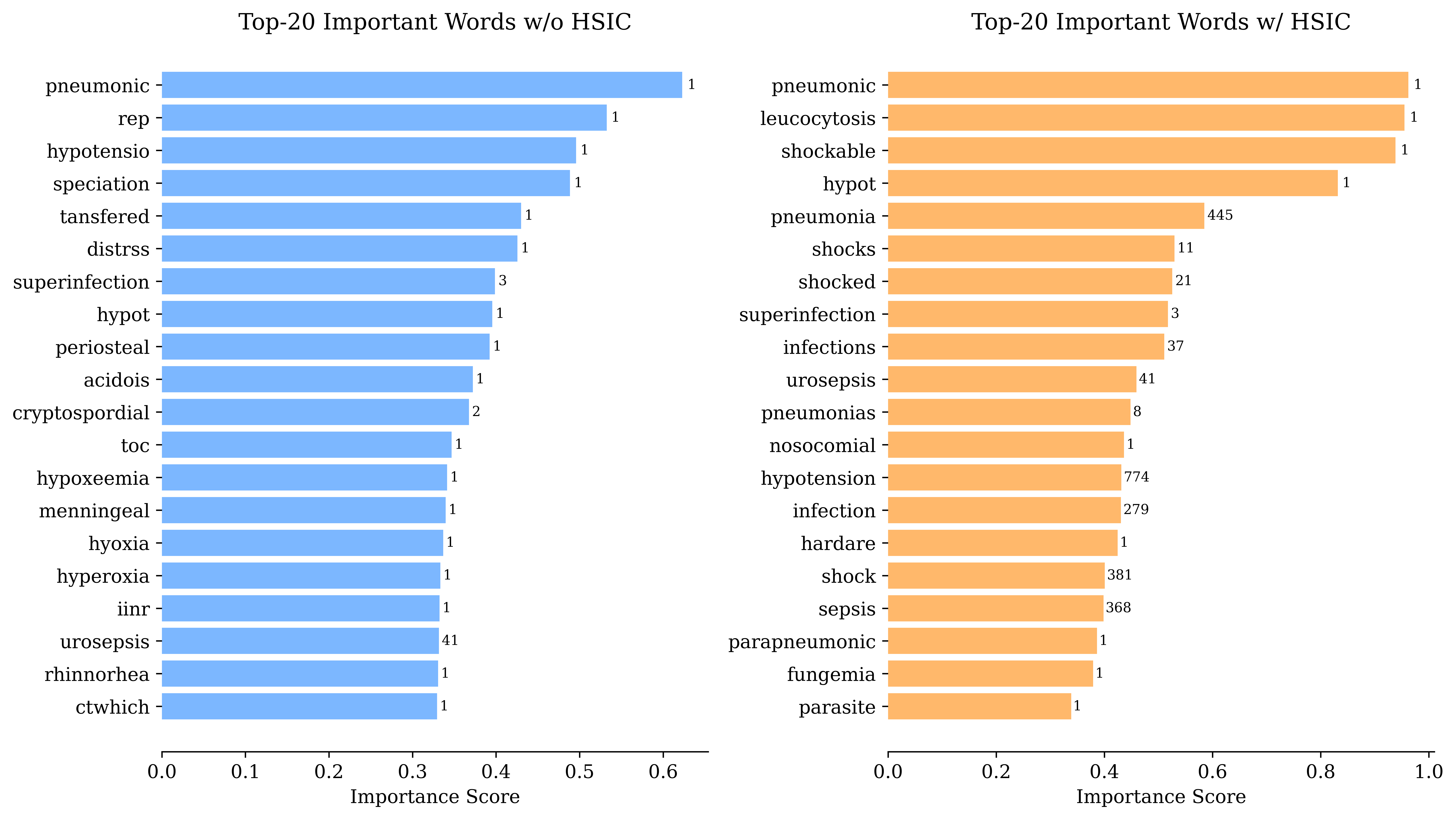}
    \caption{Comparison of importance score distributions with and without HSIC regularization.}
    \label{fig:hsic_importance}
\end{figure}

{\revvon
We quantitatively evaluate rationale quality in the main semi-synthetic setting of Section~\ref{sec:experiments}, where the confounding vocabulary is specified in the data-generating mechanism and token-level ground truth is available. 
Table~\ref{tab:rationale_quality} reports per-document average precision and recall among the top-$k$ ranked tokens for CATR and a random-ranking baseline, which shuffles CATR's token weights across valid tokens within each document.
Average precision summarizes how highly the known confounding tokens are ranked, while Recall@$k$ measures the fraction retrieved among the top $k$ tokens.

\begin{table}[!t]
\centering
\revvon
\footnotesize
\caption{\revv{Rationale quality in the main semi-synthetic setting. Metrics are averaged over documents within each seed and then over $20$ seeds.}}
\label{tab:rationale_quality}
\begin{tabular}{lrr}
\toprule
Metric & CATR & Random \\
\midrule
Per-document average precision & 0.906 & 0.048 \\
Recall@5 & 0.864 & 0.033 \\
Recall@10 & 0.960 & 0.067 \\
Recall@20 & 0.995 & 0.129 \\
\bottomrule
\end{tabular}
\end{table}

These results show that the learned token rankings concentrate on the known confounding vocabulary in this semi-synthetic setting, complementing the qualitative token visualizations in Figures~\ref{fig:mimic3} and~\ref{fig:hsic_importance}.
Figure~\ref{fig:mimic3} illustrates the selected tokens and their importance scores, while Figure~\ref{fig:hsic_importance} compares importance score distributions with and without HSIC regularization.
\par}

\section{Structured Variable List of Real Data}
\label{sec:structured_vars}
The following table lists the structured variables used for the MIMIC-III case study. All these variables are available in MIMIC-III but are joined from different tables.

\begin{table}[t]
\scriptsize
\centering
\caption{Structured variables from MIMIC-III.}
\label{tab:mimic-vars}
\renewcommand{\arraystretch}{1.1}
\begin{tabular}{llp{6.5cm}}
\toprule
\textbf{Category} & \textbf{Variable} & \textbf{Description} \\
\midrule
\multirow[t]{5}{*}{Demographics \& Baseline}
  & age & Patient age (truncated at 89) \\
  & gender & Biological sex \\
  & elixhauser & Elixhauser comorbidity score \\
  & SOFA & Sequential Organ Failure Assessment score \\
  & SIRS & Systemic Inflammatory Response Syndrome score \\
\midrule
\multirow[t]{8}{*}{Vital Signs}
  & GCS & Glasgow Coma Scale (level of consciousness) \\
  & HR & Heart rate (beats per minute) \\
  & SysBP & Systolic blood pressure \\
  & MeanBP & Mean arterial pressure \\
  & RR & Respiratory rate \\
  & SpO2 & Oxygen saturation (\%) \\
  & Temp\_C & Body temperature (°C) \\
  & Shock\_Index & Heart rate divided by systolic blood pressure \\
\midrule
\multirow[t]{3}{*}{Respiratory Status}
  & FiO2\_1 & Fraction of inspired oxygen \\
  & PaO2\_FiO2 & Oxygenation index (PaO2 / FiO2) \\
  & mechvent & Mechanical ventilation applied (baseline proxy) \\
\midrule
\multirow[t]{24}{*}{Laboratory Measurements}
  & Hb & Hemoglobin concentration \\
  & WBC\_count & White blood cell count \\
  & Platelets\_count & Platelet count \\
  & Glucose & Blood glucose level \\
  & BUN & Blood urea nitrogen \\
  & Creatinine & Kidney function indicator \\
  & CO2\_mEqL & Carbon dioxide (acid-base status) \\
  & Sodium & Serum sodium \\
  & Potassium & Serum potassium \\
  & Magnesium & Serum magnesium \\
  & Calcium & Total serum calcium \\
  & Ionised\_Ca & Ionized calcium \\
  & SGOT & AST enzyme (liver function) \\
  & SGPT & ALT enzyme (liver function) \\
  & Total\_bili & Total bilirubin \\
  & Albumin & Serum albumin \\
  & PT & Prothrombin time \\
  & PTT & Partial thromboplastin time \\
  & INR & International normalized ratio \\
  & Arterial\_pH & Arterial blood pH \\
  & paCO2 & Arterial partial pressure of CO2 \\
  & Arterial\_BE & Base excess \\
  & Arterial\_lactate & Lactate concentration \\
  & HCO3 & Bicarbonate \\
\bottomrule
\end{tabular}
\end{table}

%% file: main.bib
@article{johnson2016mimic,
  title={{MIMIC-III}, a freely accessible critical care database},
  author={Johnson, Alistair EW and Pollard, Tom J and Shen, Lu and Lehman, Li-wei H and Feng, Mengling and Ghassemi, Mohammad and Moody, Benjamin and Szolovits, Peter and Anthony Celi, Leo and Mark, Roger G},
  journal={Scientific data},
  volume={3},
  number={1},
  pages={160035},
  year={2016},
  publisher={Nature Publishing Group}
}

@article{shi2019adapting,
  title={Adapting neural networks for the estimation of treatment effects},
  author={Shi, Claudia and Blei, David and Veitch, Victor},
  journal={Advances in neural information processing systems},
  volume={32},
  year={2019}
}

@article{feder-etal-2022-causal,
  title={Causal inference in natural language processing: Estimation, prediction, interpretation and beyond},
  author={Feder, Amir and Keith, Katherine A and Manzoor, Emaad and Pryzant, Reid and Sridhar, Dhanya and Wood-Doughty, Zach and Eisenstein, Jacob and Grimmer, Justin and Reichart, Roi and Roberts, Margaret E and others},
  journal={Transactions of the Association for Computational Linguistics},
  volume={10},
  pages={1138--1158},
  year={2022},
  publisher={MIT Press, One Broadway, 12th Floor, Cambridge, Massachusetts 02142, USA}
}

@inproceedings{pmlr-v124-veitch20a,
  title={Adapting text embeddings for causal inference},
  author={Veitch, Victor and Sridhar, Dhanya and Blei, David},
  booktitle={Conference on uncertainty in artificial intelligence},
  pages={919--928},
  year={2020},
  organization={PMLR}
}

@inproceedings{wood-doughty-etal-2018-challenges,
  title={Challenges of using text classifiers for causal inference},
  author={Wood-Doughty, Zach and Shpitser, Ilya and Dredze, Mark},
  booktitle={Proceedings of the 2018 conference on empirical methods in natural language processing},
  pages={4586--4598},
  year={2018}
}

@inproceedings{lei-etal-2016-rationalizing,
  title={Rationalizing neural predictions},
  author={Lei, Tao and Barzilay, Regina and Jaakkola, Tommi},
  booktitle={Proceedings of the 2016 conference on empirical methods in natural language processing},
  pages={107--117},
  year={2016}
}

@inproceedings{bao-etal-2018-deriving,
  title={Deriving machine attention from human rationales},
  author={Bao, Yujia and Chang, Shiyu and Yu, Mo and Barzilay, Regina},
  booktitle={Proceedings of the 2018 conference on empirical methods in natural language processing},
  pages={1903--1913},
  year={2018}
}

@inproceedings{jain-etal-2020-learning,
  title={Learning to faithfully rationalize by construction},
  author={Jain, Sarthak and Wiegreffe, Sarah and Pinter, Yuval and Wallace, Byron C},
  booktitle={Proceedings of the 58th annual meeting of the association for computational linguistics},
  pages={4459--4473},
  year={2020}
}

@inproceedings{guerreiro-martins-2021-spectra,
  title={{SPECTRA}: Sparse structured text rationalization},
  author={Guerreiro, Nuno M and Martins, Andr{\'e} FT},
  booktitle={Proceedings of the 2021 Conference on Empirical Methods in Natural Language Processing},
  pages={6534--6550},
  year={2021}
}

@inproceedings{antognini2021rationalizationconcepts,
  title={Rationalization through concepts},
  author={Antognini, Diego and Faltings, Boi},
  booktitle={Findings of the Association for Computational Linguistics: ACL-IJCNLP 2021},
  pages={761--775},
  year={2021}
}

@inproceedings{yu-etal-2019-rethinking,
  title={Rethinking cooperative rationalization: Introspective extraction and complement control},
  author={Yu, Mo and Chang, Shiyu and Zhang, Yang and Jaakkola, Tommi},
  booktitle={Proceedings of the 2019 Conference on Empirical Methods in Natural Language Processing and the 9th International Joint Conference on Natural Language Processing (EMNLP-IJCNLP)},
  pages={4094--4103},
  year={2019}
}

@inproceedings{bastings-etal-2019-interpretable,
  title={Interpretable neural predictions with differentiable binary variables},
  author={Bastings, Jasmijn and Aziz, Wilker and Titov, Ivan},
  booktitle={Proceedings of the 57th Annual Meeting of the Association for Computational Linguistics},
  pages={2963--2977},
  year={2019}
}

@inproceedings{geng-etal-2020-selective,
  title={How does selective mechanism improve self-attention networks?},
  author={Geng, Xinwei and Wang, Longyue and Wang, Xing and Qin, Bing and Liu, Ting and Tu, Zhaopeng},
  booktitle={Proceedings of the 58th Annual Meeting of the Association for Computational Linguistics},
  pages={2986--2995},
  year={2020}
}

@article{Sha_Camburu_Lukasiewicz_2021,
  title={Rationalizing predictions by adversarial information calibration},
  author={Sha, Lei and Camburu, Oana-Maria and Lukasiewicz, Thomas},
  journal={Artificial Intelligence},
  volume={315},
  pages={103828},
  year={2023},
  publisher={Elsevier}
}

@article{liu2022frfoldedrationalizationunified,
  title={{FR}: Folded rationalization with a unified encoder},
  author={Liu, Wei and Wang, Haozhao and Wang, Jun and Li, Ruixuan and Yue, Chao and Zhang, YuanKai},
  journal={Advances in Neural Information Processing Systems},
  volume={35},
  pages={6954--6966},
  year={2022}
}

@article{bang2005doubly,
  title={Doubly robust estimation in missing data and causal inference models},
  author={Bang, Heejung and Robins, James M},
  journal={Biometrics},
  volume={61},
  number={4},
  pages={962--973},
  year={2005},
  publisher={Oxford University Press}
}

@inproceedings{devlin2019bertpretrainingdeepbidirectional,
  title={{BERT}: Pre-training of deep bidirectional transformers for language understanding},
  author={Devlin, Jacob and Chang, Ming-Wei and Lee, Kenton and Toutanova, Kristina},
  booktitle={Proceedings of the 2019 conference of the North American chapter of the association for computational linguistics: human language technologies, volume 1 (long and short papers)},
  pages={4171--4186},
  year={2019}
}

@article{Roberts2020AdjustingFC,
  title={Adjusting for confounding with text matching},
  author={Roberts, Margaret E and Stewart, Brandon M and Nielsen, Richard A},
  journal={American Journal of Political Science},
  volume={64},
  number={4},
  pages={887--903},
  year={2020},
  publisher={Wiley Online Library}
}

@article{egami2022make,
  title={How to make causal inferences using texts},
  author={Egami, Naoki and Fong, Christian J and Grimmer, Justin and Roberts, Margaret E and Stewart, Brandon M},
  journal={Science Advances},
  volume={8},
  number={42},
  pages={eabg2652},
  year={2022},
  publisher={American Association for the Advancement of Science}
}

@article{belloni2014high,
  title={High-dimensional methods and inference on structural and treatment effects},
  author={Belloni, Alexandre and Chernozhukov, Victor and Hansen, Christian},
  journal={Journal of Economic Perspectives},
  volume={28},
  number={2},
  pages={29--50},
  year={2014},
  publisher={American Economic Association, 2014 Broadway, Suite 305, Nashville, TN 37203~…}
}

@article{rosenbaum1983central,
  title={The central role of the propensity score in observational studies for causal effects},
  author={Rosenbaum, Paul R and Rubin, Donald B},
  journal={Biometrika},
  volume={70},
  number={1},
  pages={41--55},
  year={1983},
  publisher={Oxford University Press}
}

@article{holland1986statistics,
  title={Statistics and causal inference},
  author={Holland, Paul W},
  journal={Journal of the American statistical Association},
  volume={81},
  number={396},
  pages={945--960},
  year={1986},
  publisher={Taylor \& Francis}
}

@article{rubin1974estimating,
  title={Estimating causal effects of treatments in randomized and nonrandomized studies.},
  author={Rubin, Donald B},
  journal={Journal of educational Psychology},
  volume={66},
  number={5},
  pages={688},
  year={1974},
  publisher={American Psychological Association}
}

@article{Farrell_2021,
  title={Deep neural networks for estimation and inference},
  author={Farrell, Max H and Liang, Tengyuan and Misra, Sanjog},
  journal={Econometrica},
  volume={89},
  number={1},
  pages={181--213},
  year={2021},
  publisher={Wiley Online Library}
}

@article{zhao2023survey,
  title={A survey of large language models},
  author={Zhao, Wayne Xin and Zhou, Kun and Li, Junyi and Tang, Tianyi and Dong, Zican and Hou, Yupeng and Zhang, Beichen and Min, Yingqian and Zhang, Junjie and Liu, Peiyu and others},
  journal={Frontiers of Computer Science},
  volume={20},
  number={12},
  pages={2012627},
  year={2026},
  publisher={Springer}
}

@article{
wei2022emergent,
title={Emergent Abilities of Large Language Models},
author={Jason Wei and Yi Tay and Rishi Bommasani and Colin Raffel and Barret Zoph and Sebastian Borgeaud and Dani Yogatama and Maarten Bosma and Denny Zhou and Donald Metzler and Ed H. Chi and Tatsunori Hashimoto and Oriol Vinyals and Percy Liang and Jeff Dean and William Fedus},
journal={Transactions on Machine Learning Research},
issn={2835-8856},
year={2022},
url={https://openreview.net/forum?id=yzkSU5zdwD},
note={Survey Certification}
}

@book{Pearl-Causality,
  title={Causality: models, reasoning, and inference},
  author={Pearl, Judea},
  year={2000},
  publisher={Cambridge University Press}
}

@inproceedings{keller2015neural,
  title={Neural networks for propensity score estimation: Simulation results and recommendations},
  author={Keller, Bryan and Kim, Jee-Seon and Steiner, Peter M},
  booktitle={Quantitative Psychology Research: The 79th Annual Meeting of the Psychometric Society, Madison, Wisconsin, 2014},
  pages={279--291},
  year={2015},
  organization={Springer}
}

@article{crump2009dealing,
  title={Dealing with limited overlap in estimation of average treatment effects},
  author={Crump, Richard K and Hotz, V Joseph and Imbens, Guido W and Mitnik, Oscar A},
  journal={Biometrika},
  volume={96},
  number={1},
  pages={187--199},
  year={2009},
  publisher={Oxford University Press}
}

@inproceedings{shalit2017estimating,
  title={Estimating individual treatment effect: generalization bounds and algorithms},
  author={Shalit, Uri and Johansson, Fredrik D and Sontag, David},
  booktitle={International conference on machine learning},
  pages={3076--3085},
  year={2017},
  organization={PMLR}
}

@article{louizos2017causal,
  title={Causal effect inference with deep latent-variable models},
  author={Louizos, Christos and Shalit, Uri and Mooij, Joris M and Sontag, David and Zemel, Richard and Welling, Max},
  journal={Advances in neural information processing systems},
  volume={30},
  year={2017}
}

@inproceedings{zhang2021treatment,
  title={Treatment effect estimation with disentangled latent factors},
  author={Zhang, Weijia and Liu, Lin and Li, Jiuyong},
  booktitle={Proceedings of the AAAI Conference on Artificial Intelligence},
  volume={35},
  number={12},
  pages={10923--10930},
  year={2021}
}

@inproceedings{zhou2023causal,
  title={Causal inference from text: Unveiling interactions between variables},
  author={Zhou, Yuxiang and He, Yulan},
  booktitle={Findings of the Association for Computational Linguistics: EMNLP 2023},
  pages={10559--10571},
  year={2023}
}

@inproceedings{zhang2023towards,
  title={Towards trustworthy explanation: On causal rationalization},
  author={Zhang, Wenbo and Wu, Tong and Wang, Yunlong and Cai, Yong and Cai, Hengrui},
  booktitle={International Conference on Machine Learning},
  pages={41715--41736},
  year={2023},
  organization={PMLR}
}

@misc{ebrahimi2023lanistr,
      title={{LANISTR}: Multimodal Learning from Structured and Unstructured Data}, 
      author={Sayna Ebrahimi and Sercan O. Arik and Yihe Dong and Tomas Pfister},
      year={2024},
      eprint={2305.16556},
      archivePrefix={arXiv},
      primaryClass={cs.LG},
      url={https://arxiv.org/abs/2305.16556}, 
}

@article{cole2008constructing,
  title={Constructing inverse probability weights for marginal structural models},
  author={Cole, Stephen R and Hern{\'a}n, Miguel A},
  journal={American journal of epidemiology},
  volume={168},
  number={6},
  pages={656--664},
  year={2008},
  publisher={Oxford University Press}
}

@article{lee2011weight,
  title={Weight trimming and propensity score weighting},
  author={Lee, Brian K and Lessler, Justin and Stuart, Elizabeth A},
  journal={PloS one},
  volume={6},
  number={3},
  pages={e18174},
  year={2011},
  publisher={Public Library of Science, San Francisco, USA}
}

@article{li2018balancing,
  title={Balancing covariates via propensity score weighting},
  author={Li, Fan and Morgan, Kari Lock and Zaslavsky, Alan M},
  journal={Journal of the American Statistical Association},
  volume={113},
  number={521},
  pages={390--400},
  year={2018},
  publisher={Taylor \& Francis}
}

@article{imai2014covariate,
  title={Covariate balancing propensity score},
  author={Imai, Kosuke and Ratkovic, Marc},
  journal={Journal of the Royal Statistical Society Series B: Statistical Methodology},
  volume={76},
  number={1},
  pages={243--263},
  year={2014},
  publisher={Oxford University Press}
}

@article{van2025stabilized,
  title={Stabilized inverse probability weighting via isotonic calibration},
  author={van der Laan, Lars and Lin, Ziming and Carone, Marco and Luedtke, Alex},
  journal={Proceedings of machine learning research},
  volume={275},
  pages={139},
  year={2025}
}

@article{gretton2007kernel,
  title={A kernel statistical test of independence},
  author={Gretton, Arthur and Fukumizu, Kenji and Teo, Choon and Song, Le and Sch{\"o}lkopf, Bernhard and Smola, Alex},
  journal={Advances in neural information processing systems},
  volume={20},
  year={2007}
}

@article{cai2023learning,
  title={On learning necessary and sufficient causal graphs},
  author={Cai, Hengrui and Wang, Yixin and Jordan, Michael and Song, Rui},
  journal={Advances in Neural Information Processing Systems},
  volume={36},
  pages={42148--42160},
  year={2023}
}

@inproceedings{gretton2005measuring,
  title={Measuring statistical dependence with {Hilbert-Schmidt} norms},
  author={Gretton, Arthur and Bousquet, Olivier and Smola, Alex and Sch{\"o}lkopf, Bernhard},
  booktitle={International conference on algorithmic learning theory},
  pages={63--77},
  year={2005},
  organization={Springer}
}

@article{song2012feature,
  author  = {Le Song and Alex Smola and Arthur Gretton and Justin Bedo and Karsten Borgwardt},
  title   = {Feature Selection via Dependence Maximization},
  journal = {Journal of Machine Learning Research},
  year    = {2012},
  volume  = {13},
  number  = {47},
  pages   = {1393--1434},
  url     = {http://jmlr.org/papers/v13/song12a.html}
}

@inproceedings{zhang2011kernel,
  title={Kernel-based Conditional Independence Test and Application in Causal Discovery},
  author={Zhang, Kun and Peters, Jonas and Janzing, Dominik and Schoelkopf, Bernhard},
  booktitle={Uncertainty in Artificial Intelligence},
  pages={885--894},
  year={2011},
  organization={PMLR}
}

@article{strobl2019approximate,
  title={Approximate kernel-based conditional independence tests for fast non-parametric causal discovery},
  author={Strobl, Eric V and Zhang, Kun and Visweswaran, Shyam},
  journal={Journal of Causal Inference},
  volume={7},
  number={1},
  pages={20180017},
  year={2019},
  publisher={De Gruyter}
}

@inproceedings{karlsson2025falsification,
  title={Falsification of Unconfoundedness by Testing Independence of Causal Mechanisms},
  author={Karlsson, Rickard and Krijthe, Jh},
  booktitle={International Conference on Machine Learning},
  pages={29128--29147},
  year={2025},
  organization={PMLR}
}

@inproceedings{pearl1995testability,
author = {Pearl, Judea},
title = {On the testability of causal models with latent and instrumental variables},
year = {1995},
isbn = {1558603859},
publisher = {Morgan Kaufmann Publishers Inc.},
address = {San Francisco, CA, USA},
booktitle = {Proceedings of the Eleventh Conference on Uncertainty in Artificial Intelligence},
pages = {435–443},
numpages = {9},
location = {Montr{\'e}al, Qu{\'e}, Canada},
series = {UAI'95}
}

@article{wang2023optimized,
  title={Optimized glycemic control of type 2 diabetes with reinforcement learning: a proof-of-concept trial},
  author={Wang, Guangyu and Liu, Xiaohong and Ying, Zhen and Yang, Guoxing and Chen, Zhiwei and Liu, Zhiwen and Zhang, Min and Yan, Hongmei and Lu, Yuxing and Gao, Yuanxu and others},
  journal={Nature medicine},
  volume={29},
  number={10},
  pages={2633--2642},
  year={2023},
  publisher={Nature Publishing Group US, New York}
}

@article{liu2025generalist,
  title={A generalist medical language model for disease diagnosis assistance},
  author={Liu, Xiaohong and Liu, Hao and Yang, Guoxing and Jiang, Zeyu and Cui, Shuguang and Zhang, Zhaoze and Wang, Huan and Tao, Liyuan and Sun, Yongchang and Song, Zhu and others},
  journal={Nature medicine},
  volume={31},
  number={3},
  pages={932--942},
  year={2025},
  publisher={Nature Publishing Group US, New York}
}

@article{Hoyer2009,
  title={Nonlinear causal discovery with additive noise models},
  author={Hoyer, Patrik and Janzing, Dominik and Mooij, Joris M and Peters, Jonas and Sch{\"o}lkopf, Bernhard},
  journal={Advances in neural information processing systems},
  volume={21},
  year={2008}
}

@article{Peters2014causal,
  author  = {Jonas Peters and Joris M. Mooij and Dominik Janzing and Bernhard Sch{{\"o}}lkopf},
  title   = {Causal Discovery with Continuous Additive Noise Models},
  journal = {Journal of Machine Learning Research},
  year    = {2014},
  volume  = {15},
  number  = {58},
  pages   = {2009--2053},
  url     = {http://jmlr.org/papers/v15/peters14a.html}
}

@inproceedings{
gui2023causal,
title={Causal Estimation for Text Data with (Apparent) Overlap Violations},
author={Lin Gui and Victor Veitch},
booktitle={The Eleventh International Conference on Learning Representations },
year={2023},
url={https://openreview.net/forum?id=Ha2MnQM9Ph}
}

@inproceedings{
melnychuk2024bounds,
title={Bounds on Representation-Induced Confounding Bias for Treatment Effect Estimation},
author={Valentyn Melnychuk and Dennis Frauen and Stefan Feuerriegel},
booktitle={The Twelfth International Conference on Learning Representations},
year={2024},
url={https://openreview.net/forum?id=d3xKPQVjSc}
}

@inproceedings{shiyu2020invariant,
  title={Invariant rationalization},
  author={Chang, Shiyu and Zhang, Yang and Yu, Mo and Jaakkola, Tommi},
  booktitle={International Conference on Machine Learning},
  pages={1448--1458},
  year={2020},
  organization={PMLR}
}
